\documentclass{article}

    \PassOptionsToPackage{numbers, compress}{natbib}

    \usepackage[final]{want_neurips_2023}

\usepackage[utf8]{inputenc} 
\usepackage[T1]{fontenc}    
\usepackage{hyperref}       
\usepackage{url}            
\usepackage{booktabs} 
\usepackage{amsmath}
\usepackage{amsfonts}       
\usepackage{nicefrac}       
\usepackage{microtype}      
\usepackage{xcolor}         
\usepackage{graphicx}
\usepackage{subcaption}
\usepackage{wrapfig}
\usepackage[most]{tcolorbox}
\usepackage{bbm}
\usepackage{multirow}
\usepackage{algorithm}
\usepackage{listings}

\newcommand{\params}{\boldsymbol{\theta}}
\newcommand{\method}{LAWA}
\newcommand{\lawasol}{\params^{\text{\method}}}
\DeclareUnicodeCharacter{03A3}{\ensuremath{\Sigma}}
\definecolor{lightgray}{gray}{0.9}

\title{Early Weight Averaging meets High Learning Rates for LLM Pre-training}

\author{%
  \textbf{Sunny Sanyal}\thanks{These authors contributed equally to this work. Correspondence to sanyal.sunny@utexas.edu} 
  \vspace{0.25em} \\
  UT Austin \\
  \And
  \textbf{Atula Neerkaje}\footnotemark[1]
  \vspace{0.25em} \\
  UT Austin \\
  \And
  \textbf{Jean Kaddour} 
  \vspace{0.25em} \\
  UCL \\
  \AND
  \textbf{Abhishek Kumar} 
  \vspace{0.25em} \\
  Google DeepMind \\
  \And
  \textbf{Sujay Sanghavi} 
  \vspace{0.25em} \\
  UT Austin \\
}
\begin{document}

\maketitle

\begin{abstract}
 Training Large Language Models (LLMs) incurs significant cost; hence, any strategy that accelerates model convergence is helpful. In this paper, we investigate the ability of a simple idea – checkpoint averaging along the trajectory of a training run – to improve both convergence and generalization quite early during training. Here we show that models trained with high learning rates observe higher gains due to checkpoint averaging. Furthermore, these gains are amplified when checkpoints are sampled with considerable spacing in training steps. Our training recipe outperforms conventional training and popular checkpoint averaging baselines such as exponential moving average (EMA) and stochastic moving average (SWA). We evaluate our training recipe by pre-training LLMs, where high learning rates are inherently preferred due to extremely large batch sizes. Specifically, we pre-trained nanoGPT-2 models of varying sizes—small (125M), medium (335M), and large (770M)—on the OpenWebText dataset, comprised of 9B tokens. Additionally, we present results for publicly available Pythia LLMs, ranging from 1B to 12B, which were trained on the PILE-deduped dataset containing 207B tokens. Code is available at \url{https://github.com/sanyalsunny111/Early_Weight_Avg}.
\end{abstract}

\section{Introduction} \label{sec:intro}

Large Language Models (LLMs) have made a significant leap from billion to trillion scale, both in terms of parameters \citep{chowdhery2022palm, ren2023pangu} and pre-training data size \citep{chinchilla} \cite{touvron2023llama, touvron2023llama2}. This surge in both data and model size has rendered LLM pre-training increasingly slow and resource-intensive. For instance, a Llama 2 70B model trained with 2T tokens took 1720K GPU hours to train. To accelerate the training process, it is a popular practice in LLM pre-training \citep{biderman2023pythia, touvron2023llama} to utilize exceptionally large batch sizes, thereby ensuring maximal GPU utilization. The usage of large batch sizes requires scaling the learning rates proportional to it's batchsize \citep{goyal2017accurate}, \citep{krizhevsky2014one} for SGD or proportional to the square root of it's batch size for adaptive gradient methods \citep{malladi2022sdes}. Overall, high learning rates are preferred when utilizing large batch sizes.

In this paper, our goal is to improve the test generalization (log perplexity) of LLM pre-training while reducing the number of training steps, all without increasing the compute budget. To achieve this, we first demonstrate that: (a) models trained with higher learning rates exhibit greater improvements when averaged along the training trajectory, and (b) averaging distant model weights from a single training trajectory further amplifies these gains. We integrate these two insights to adapt LAWA (LAtest Weight Averaging) \citep{lawa}—a technique that performs checkpoint averaging throughout a training trajectory using a sliding window—for pre-training LLMs.

We evaluate our methodology by pre-training nanoGPT-2 models of various scales, specifically 125M (small), 355M (medium), and 770M (large), using the OpenWebText dataset, which comprises 9B tokens. The experiments with nanoGPT-2 are conducted in a controlled environment to gain a deeper understanding of our training recipe. Furthermore, we extend our evaluation to publicly available Pythia LLMs \citep{biderman2023pythia}, which include model sizes of 1B, 2.8B, 6.9B, and 12B, trained using 207B tokens. Our experiments with Pythia LLMs aim to demonstrate the impact of our work on real-world LLMs. 

\begin{figure}[t]
  \begin{subfigure}{0.33\textwidth} 
    \centering
    \includegraphics[width=\linewidth]{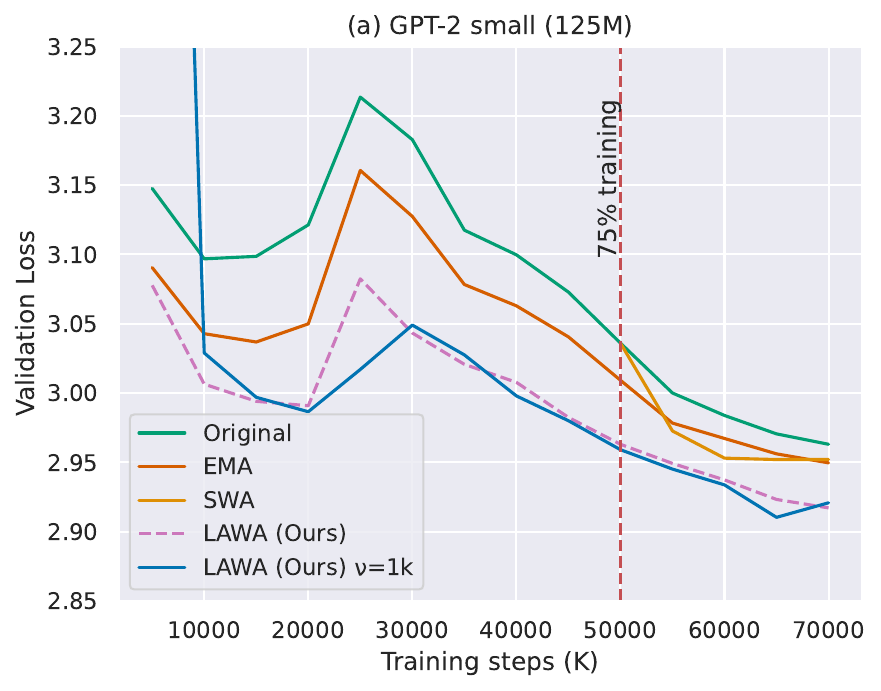} 
    \caption{}
  \end{subfigure}
  \begin{subfigure}{0.33\textwidth}
    \centering
    \includegraphics[width=\linewidth]{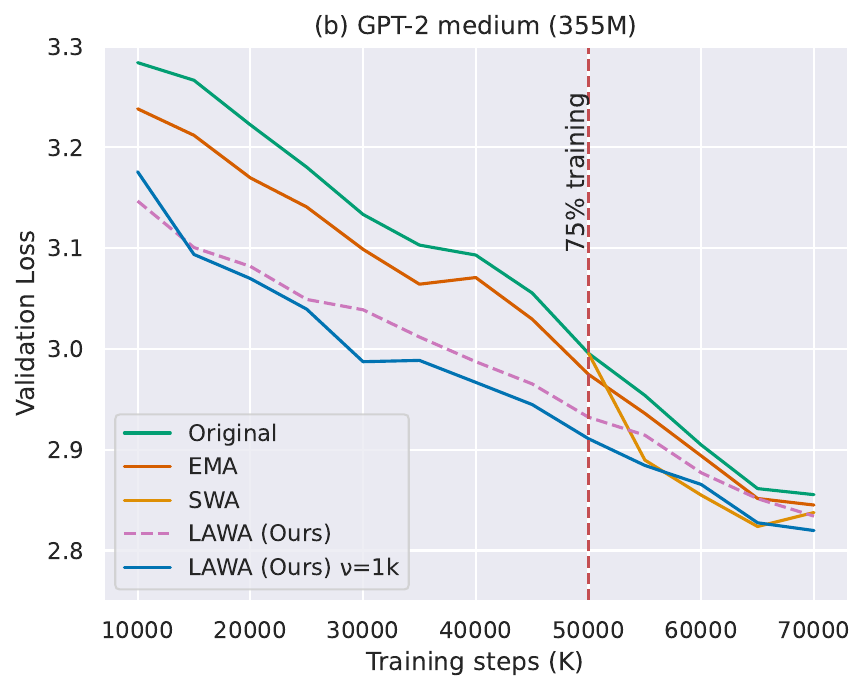} 
    \caption{}
  \end{subfigure}
  \begin{subfigure}{0.33\textwidth}
    \centering
    \includegraphics[width=\linewidth]{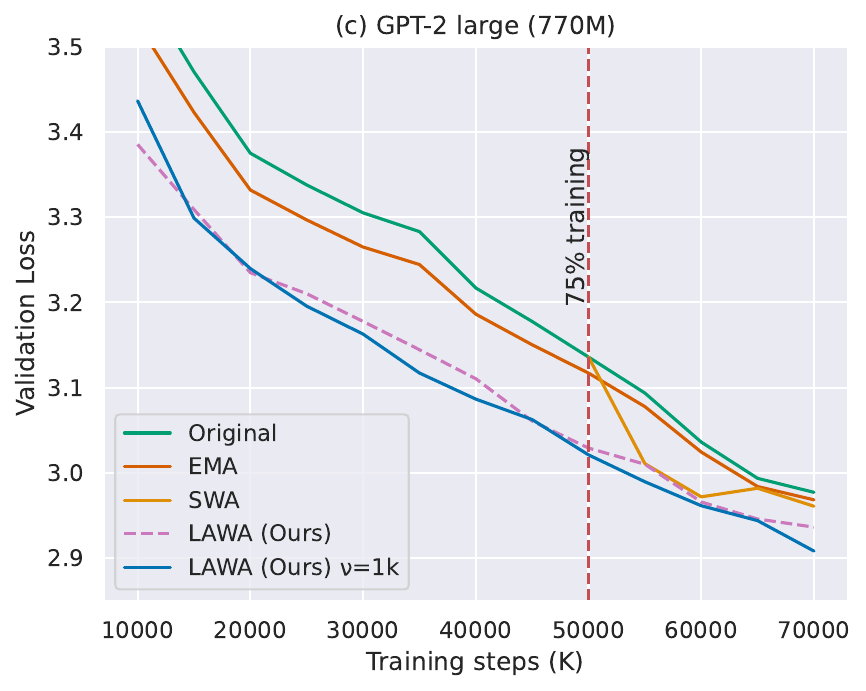} 
    \caption{}
  \end{subfigure}
  \caption{Across all model sizes, LAWA achieves faster convergence and generalizes better in comparison to original pretraining run and other baseline averaging schemes. Validation loss on OpenWebText with 70K training steps; (a) GPT2-small (125M) with Original is 2.963, EMA is 2.949, SWA is 2.952 and LAWA (ours-best) is 2.917, (b) GPT2-medium (355M) with Original is 2.855, EMA is 2.845, SWA is 2.837 and LAWA (ours-best) is 2.819, and (c) GPT2-large (770M) with Original is 2.977, EMA is 2.968, SWA is 2.961 and LAWA (ours-best) is 2.908.}
  \label{fig:nanogpt2_main}
\end{figure}

\begin{figure}[t]
  \begin{subfigure}{0.33\textwidth} 
    \centering
    \includegraphics[width=\linewidth]{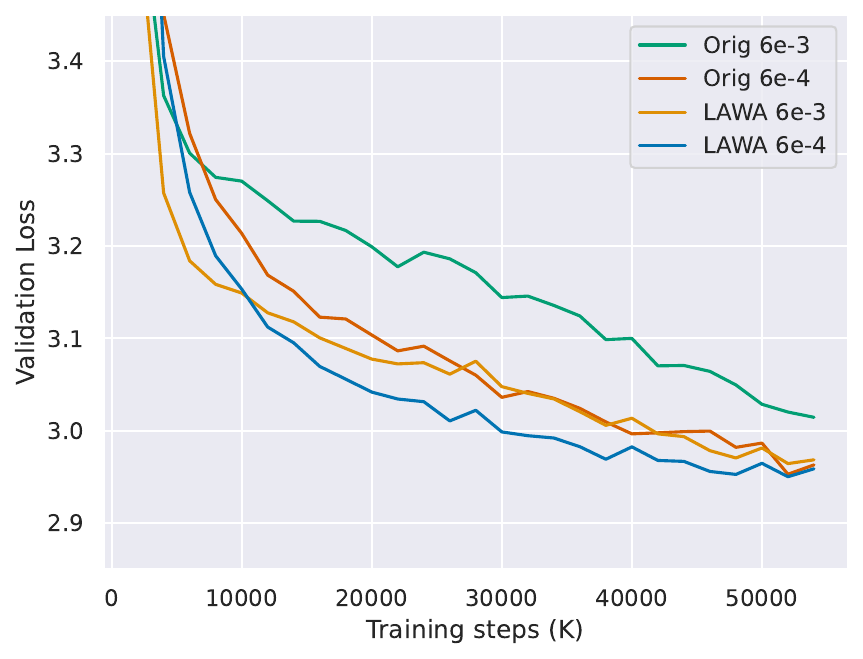} 
    \caption{}
  \end{subfigure}
  \begin{subfigure}{0.33\textwidth}
    \centering
    \includegraphics[width=\linewidth]{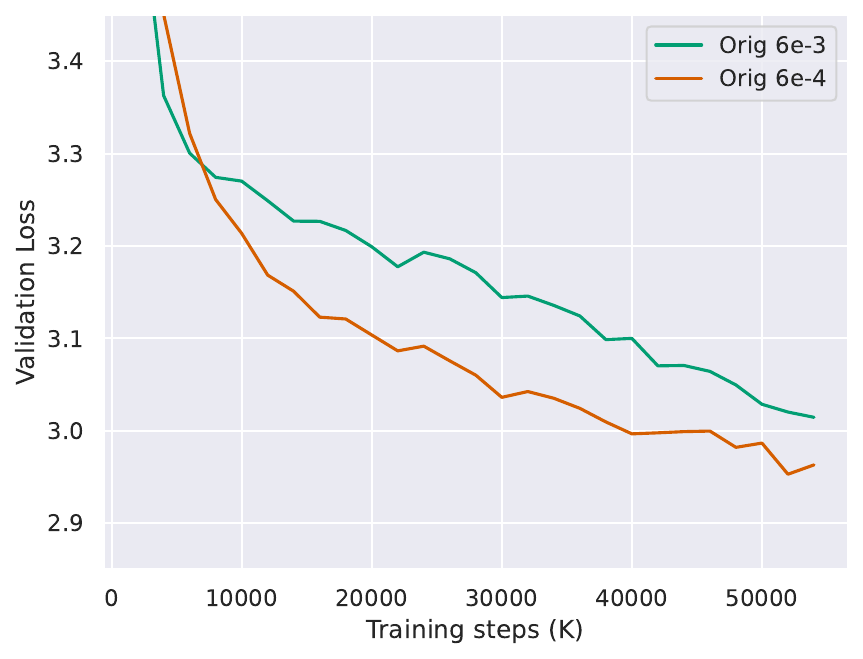} 
    \caption{}
  \end{subfigure}
  \begin{subfigure}{0.33\textwidth}
    \centering
    \includegraphics[width=\linewidth]{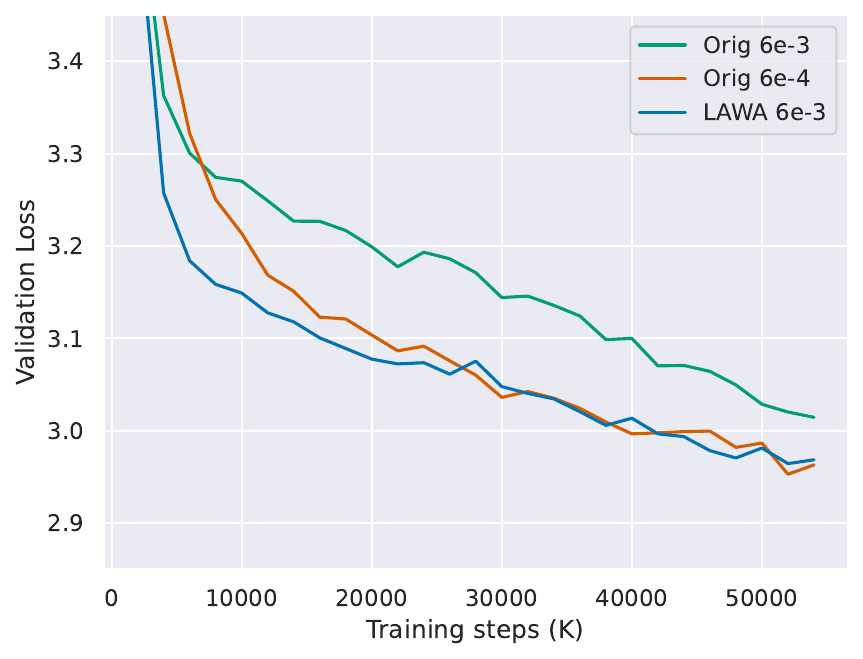} 
    \caption{}
  \end{subfigure}
  \caption{We compare two independently trained nanoGPT-2 (125M) models with LR $=[\mathsf{6 \times 10^{-3}}, \mathsf{6 \times 10^{-4}}]$ on OpenWebText data. (a) Pre-training curve with and without LAWA. LLMs trained with higher LR observes higher gain due to \method{}. (b) The model trained with a high LR generalizes poorly compared to its counterpart trained with low/tuned LR. (c) The generalization gap caused by the high LR is effectively mitigated by \method{}.}
  \label{fig:nanogpt2_lr}
\end{figure}

\paragraph{Main Contributions.} In summary, our findings are as follows,

\begin{enumerate}

\item We empirically show that models trained with high learning rate (LR) show pronounced gains over original training on performing checkpoint averaging very early on during training (Figure \ref{fig:nanogpt2_lr}). This gain further amplifies when we sample distant checkpoints in the training run (Figure \ref{fig:nanogpt2_main}). We provide a intuitive explanation of this phenomenon in Section \ref{sec:method}.

\item We observe that the training trajectory of \method{} closely mimics that of a model being trained with a low LR. The primary advantage of \method{} is that it allows LLMs to be trained with high LR without compromising generalization (Section \ref{sec:results}).

\item We show that \method{} improves test generalization with fewer training steps compared to original training starting very early on during training; for both nanoGPT-2 and Pythia LLMs (Figure \ref{fig:nanogpt2_main} and Figures \ref{fig:pile1B}-\ref{fig:pile12B}). \method{} also improves zero-shot performance for nanoGPT-2 and Pythia LLMs, as shown in Table \ref{table:nanogpt2_zeroshot}, \ref{table:zeroshot}.

\item We compare our recipe with conventional training and popular baselines such as Exponential Moving Average - EMA \citep{szegedy2015rethinking} and Stochastic Weight Averaging - SWA \citep{SWA}. These baselines were not originally proposed or evaluated for LLM pre-training, but we adapt them to set meaningful baselines. Our training recipe outperforms conventional pre-training, EMA, and SWA.

\item Additionally, we perform a preliminary investigation of early weight averaging for a diffusion model for image generation (specifically, a 422M sized UNet model trained with the standard DDPM objective \citep{ho2020denoising}). We observe thematically similar improvements (evaluated by the FID metric) as shown in Figure \ref{fig:fid_diffusion}. 

\end{enumerate}

\paragraph{Paper outline.} The structure of the remainder of this paper is as follows: Initially, we introduce the problem with a simple example and explain the intuition behind our approach in Section \ref{sec:method}. This is followed by a detailed description of our experimental setup in Section \ref{sec:exp}, and the presentation of our main findings in Section \ref{sec:results}. Subsequently, we compare our work with previous studies in Section \ref{sec:rw}. The paper concludes with a summary of our key findings and suggests possible avenues for future research.

\section{Intuition and Method} \label{sec:method}

\subsection{Toy Setting}

\begin{wrapfigure}{r}{0.5\textwidth} 
\vspace{-2ex}
  \centering
  \includegraphics[width=0.4\textwidth]{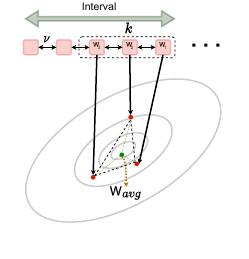} 
  \caption{LAWA illustration: Given weights $\mathsf{W}_1, \mathsf{W}_2, ... \mathsf{W_k}$ from a high LR trajectory separated by k-stepsize ($\nu$, Algorithm \ref{alg:lawa_algo}), LAWA computes $\mathsf{W}_{avg}$ at a given step.\protect\footnotemark}
  \label{fig:lawa_toy}
\end{wrapfigure}
\footnotetext{Note that $\mathsf{W}$ refers to the set of weights $\{w_1, w_2, ...\}$.}

We explain the setting using a simple toy problem of minimizing a two-dimensional loss function, represented as \( L(w_{1}, w_{2}) \), where \( w_{1} \) and \( w_{2} \) are the parameters of the model. In this scenario, there exists an optimal batch size, \( \mathcal{B}_{0} \), and an optimal learning rate, \( \eta_{0} \), that minimizes the loss function. Assuming, for reasons outlined in Section \ref{sec:intro}, we are compelled to use a batch size, \( \mathcal{B} \), and a learning rate, \( \eta \), such that both are significantly larger than their optimal counterparts, i.e., \( \mathcal{B} \gg \mathcal{B}_{0} \) and \( \eta \gg \eta_{0} \) respectively. Suppose also that the loss function \( L(w_{1}, w_{2}) \) exhibits a much higher curvature along \( w_{1} \), as compared to \( w_{2} \). It is widely known that the updates from the AdamW optimizer are mostly uniform across all weight dimensions \citep{liu2023sophia}. When \( \eta \gg \eta_{0} \) the weight updates of \( w_{2} \) will be accelerated, however this will cause oscillations along \( w_{1} \) which in a long run adversely affects the convergence of \( w_{1} \). A naive approach to mitigate this problem is to use a smaller LR or decay LR to 0, which might hinder progress in flatter regions. It is conceivable that a LLM might exhibit extremely heterogeneous curvatures exacerbating this issue during pre-training.

\subsection{Intuition of Our Approach}

\paragraph{Optimization Viewpoint.} We propose performing checkpoint averaging of model weights relatively early during training with high learning rates (\( \eta \)). The rationale behind this step stems from the fact that checkpoint averaging serves as a surrogate to LR decay, as demonstrated by~\citet{sandler2023training}. However, this surrogate LR decay is decoupled from the weight update during optimization, as checkpoint averaging is conducted in a post-hoc manner. Employing this simple technique, we mitigate the oscillations in \( w_{1} \) while swiftly traversing through \( w_{2} \), achieving enhanced generalization in fewer training steps as illustrated in Figure ~\ref{fig:lawa_toy}.

\paragraph{Diversity Viewpoint.} The practice of averaging the weights of model checkpoints is broadly recognized as being functionally analogous to ensembling \citep{SWA, model_soups}. In model ensembling literature it is well established that diverse models improve the performance of the ensemble \citep{lakshminarayanan2016simple}. Therefore, it is fair to assume that this principle also applies to model averaging as well. In our context, we define the diversity of a model at two distinct training steps, $\frac{1}{N} \sum_{i=1}^{N} \mathbf{1}\left[y_i^{\mathsf{W}_1} \neq y_i^{\mathsf{W}_2}\right]$ which calculates the number of disagreements between the two checkpoints. This equation computes the number of samples from the same held-out set where the checkpoint \(\mathsf{W}_1\) disagrees with checkpoint \(\mathsf{W}_2\). A recent study by \citeauthor{athiwaratkun2018many} has demonstrated that a higher LR can result in the generation of diverse model checkpoints. We observed (Figure \ref{fig:nanogpt2_main}) that this phenomenon can be further amplified by sampling far apart checkpoints in terms of training step. We combine both these insights to induce diversity in our checkpoints.

\subsection{LAWA: LAtest Weight Averaging}

We explain the Latest Weight Averaging (\method{}) algorithm below along with a python-style pseudo code (Algorithm \ref{alg:lawa_algo}). As shown in Figure \ref{fig:lawa_toy}, \method{} maintains a first in first out (FIFO) queue of periodically sampled checkpoints with a large number of intervening steps ($\nu$) in between two succesive samples. We adapt \method{} for our setting with minor modifications. Specifically, we introduce $\mathsf{k\_stepsize}$ ($\nu$), and decoupled $\mathsf{interval}$ and $\mathsf{k}$ to effectively sample distant checkpoints in the training run. The original LAWA algorithm \citep{lawa} assumes $\mathsf{interval}$ = $\mathsf{k}$.

\begin{tcolorbox}[colback=blue!5,colframe=blue!75!black]
\method{} runs a moving window at a predetermined interval to collect $\mathsf{k}$ latest checkpoints on sequence of saved checkpoints $\params_t$. The \method{} derived checkpoints are computed as $\lawasol_t := \frac{1}{\mathsf{k}} \sum_{s=t-\mathsf{k}}^{t} \params_s$ where $\params_t$ the original checkpoints are sampled several training steps apart in the training process.
\end{tcolorbox}

\section{Experimental Setup} \label{sec:exp}

\begin{wrapfigure}{r}{0.55\textwidth} 
\vspace{-6ex}
\begin{center}
\begin{minipage}{0.55\textwidth} 
\begin{algorithm}[H]
\caption{LAWA: Pytorch-style pseudocode}
\label{alg:lawa_algo}
\lstset{
  language=Python,
  backgroundcolor=\color{white},
  basicstyle=\fontsize{4.5pt}{4.5pt}\ttfamily\selectfont, 
  columns=fullflexible,
  breaklines=true,
  captionpos=b,
  commentstyle=\fontsize{4.5pt}{4.5pt}\color{gray},
  keywordstyle=\fontsize{4.5pt}{4.5pt}\color{codeblue},
}
\begin{lstlisting}[language=Python, backgroundcolor=\color{black!5}, basicstyle=\small\ttfamily, keywordstyle=\color{blue}\ttfamily, commentstyle=\color{green!40!black}\ttfamily]
def LAWA(ckpts, interval, k, k_stepsize):
    # ckpts: list of checkpoints
    # k: number of checkpoints to average
    # k_stepsize: distance in training steps between checkpoints
    averaged_ckpts = []
    for i in range(0, len(ckpts), interval):
        start_idx = i - k 
        end_idx = i - 1
        # select last k checkpoints 
        # with k_stepsize between each checkpoint.
        selected_ckpts = ckpts[end_idx-k+1:end_idx+1:k_stepsize]
        avg = average(selected_ckpts)
        averaged_ckpts.append(avg)
    return averaged_ckpts
\end{lstlisting}
\label{alg:lawa}
\end{algorithm}
\end{minipage}
\vspace{-2ex}
\end{center}
\end{wrapfigure}

\paragraph{NanoGPT-2 Experiments.} We conduct all our experiments utilizing autoregressive decoder-style Large Language Models (LLMs), specifically nanoGPT-2 and Pythia LLMs. We utilize three distinct sizes of nanoGPT-2 models: small (125M), medium (355M), and large (770M). We train nanoGPT-2 models from scratch using the OpenWebText dataset, which includes 9 billion training tokens and 4.4 million validation tokens. Throughout the experiments, we maintain a consistent sequence length of 1024 and a fixed batch size of 131K tokens per batch, the latter being the maximum batch size accommodated by our GPUs. The configurations for the model and pre-training were adapted from Sophia's \citep{liu2023sophia} AdamW baseline, with adjustments made to the learning rate and batch size to align with our specific needs. Notably, we trained all the models with learning rates that were ten times higher and batch sizes that were twice as large compared to the configurations in \cite{liu2023sophia}, where the learning rate was tuned through a grid search. We compare \method{} with the original pre-training recipe, EMA \citep{szegedy2015rethinking}, and SWA \citep{SWA}, which we adapt for LLMs. For EMA, we set the decay to 0.9 as per \cite{lawa} and update the EMA model at every step, which is a standard practice. For SWA, we adhere to the original pre-training procedure until 75\% completion, after which SWA training is initiated with a new SWA scheduler (cosine annealing). We compute the SWA uniform average every 10 steps.

\paragraph{Pythia Experiments.}The \textbf{Pythia LLMs} are publicly available in the Pythia suite \citep{biderman2023pythia}. We report results on Pythia-1B, Pythia-2.8B, Pythia-6.9B, and Pythia-12B; Table.\ref{table:interp} summarizes the details of these models. For our experiments, we use intermediate model checkpoints; saved after every 1000 update steps. The models are trained by \citet{biderman2023pythia} on the PILE dataset \citep{gao2020pile}, a publicly available, curated collection of English text corpus of size 800GB. The original PILE dataset is curated using 5 different genres of data namely, academia, internet, prose, dialogue and miscellaneous. The PILE dataset contains 300 billion tokens prior to deduplication, and this number reduces to 207 billion tokens after the deduplication process. Our experiments use Pythia models trained with PILE-deduped dataset, as such models tend to memorize less \citep{lee2021deduplicating}. The batch size for all the Pythia models was set at 2.09 million tokens and the learning rate was scaled following \cite{zhang2022opt}.

\begin{table}[t]
    \centering
    \small
    \begin{tabular}{rcccccc}\toprule
  Model Size & Layers & Hidden Size & Heads & Learning Rate        & Equivalent Models\\\midrule
         125M     & 12     & 12        & 768   & $6.0\times 10^{-3}$ & nanoGPT-2 (small) \\
         335M     & 24     & 1024      & 16    & $3.0\times 10^{-3}$ & nanoGPT-2 (medium)\\
         770 M     & 36     & 1280     & 20    & $2.0\times 10^{-3}$ & nanoGPT-2 (large) \\
         1.0 B     & 16     & 2048     &  8    & $3.0\times 10^{-4}$  & --- \\
         2.8 B     & 32     & 2560     & 32    & $1.6\times 10^{-4}$  & GPT-Neo 2.7B, OPT-2.7B\\
         6.9 B     & 32     & 4096      & 32    & $1.2\times 10^{-4}$  & OPT-6.7B\\
          12 B     & 36     & 5120      & 40    & $1.2\times 10^{-4}$  & ---\\\bottomrule
	\end{tabular}
	\vspace{0.5em}
	\caption{Overview of models and their architecture from the nanoGPT-2 suite and Pythia suite \citep{biderman2023pythia} used in our experiments. The model nomenclature for Pythia LLMs is pythia-xx with model size. Models marked as ``equivalent'' have the same architecture and number of non-embedding parameters.}
	\label{table:interp}
\end{table}

\begin{wrapfigure}{r}{0.5\textwidth}
    \vspace{-0.5cm}
    \centering
    \includegraphics[width=0.4\textwidth]{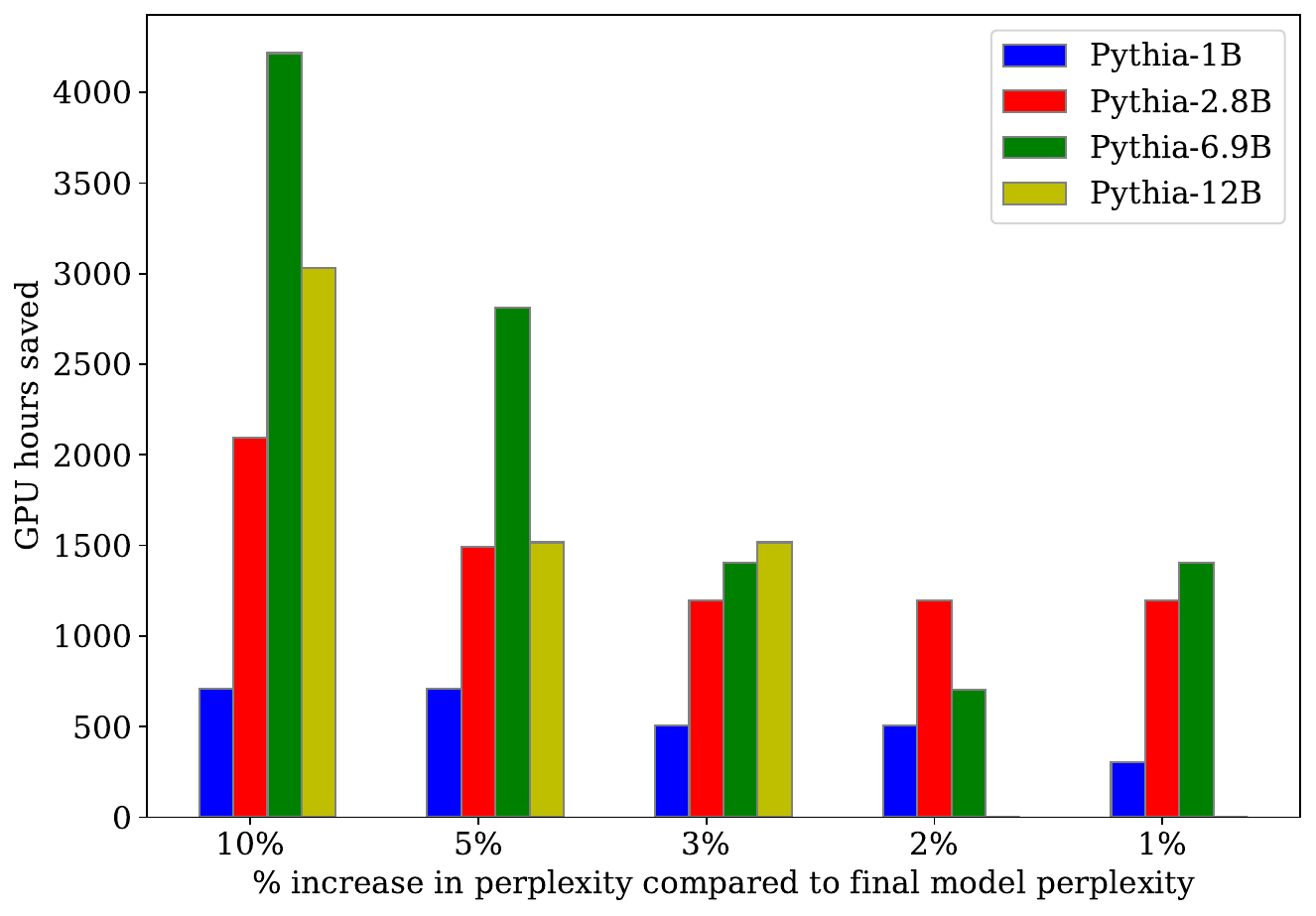}
    \caption{\method{} saves significant amount of GPU hours compared to original training. We compare the savings in GPU hours as a function of increase in final perplexity, i.e. perplexity achieved at 141K training step by the original checkpoint. This plot is created using a held out set from the training subset PILE-philosophy papers.}
    \label{fig:savings}
\end{wrapfigure}

\paragraph{Evaluation} We evaluate the language modelling performance of nanoGPT-2 models pre-trained for 70K steps using log perplexity (perplexity and loss used interchangeably) on the held-out/val set. For nanoGPT-2 we use the moving window $\mathsf{interval}=\text{1K}$, $\mathsf{k}$ = 5 and we sample checkpoints $\mathsf{interval}=\text{200,1K}$ apart for \method{}. Next we analyze the original training trajectories of Pythia LLMs and demonstrate the improvements achieved in test generalization using $\mathsf{LAWA}$. We also present zero-shot evaluation results on Lambada OpenAI \citep{paperno2016lambada}, SciQ \citep{sciq}, AI2 Reasoning Challenge-easy (ARC-e) \citep{Clark2018ThinkYH}, and Wikitext \citep{wikitext}. We evaluate 4 Pythia LLMs using the intermediate model checkpoints on a subset of the test and validation dataset following the methodology prescribed by \cite{xia2022training}. For the purpose of evaluation we use the open source library \texttt{lm-evaluation harness}\footnote{https://github.com/EleutherAI/lm-evaluation-harness}. We select representative subsets from the diverse genres encompassed by the full PILE validation and test dataset. This subset comprises PILE-philosophy papers, PILE-bookcorpus2, and PILE-YouTube subtitles datasets.

We conduct zero-shot evaluation on the Pythia LLMs. For zero-shot evaluation we provide a natural language description of the downstream task, along with a textual example. The models then generate responses that are either open-ended or discriminatively select a proposed answer. This evaluation setup serves as a robust academic benchmark, as it assesses Pythia models of various scales on reasonably large PILE subsets and downstream datasets, both in terms of test performance and zero-shot. For both the test generalization and zero-shot experiments, we evaluate model checkpoints starting 21K steps to 141K steps (recall that subsequent checkpoints are 1K steps apart). Moreover, we choose to slide the averaging window at 3K steps (i.e. $\mathsf{interval}=\text{3K}$) and average last $\mathsf{k}$ intermediate checkpoints as discussed in \method{} algorithm (Algorithm \ref{alg:lawa_algo}). Our selection of \method{} parameters such as $\mathsf{k}$ = 5 and start step = 21K are based on the experiments discussed in 
Section \ref{sec:abl}. 

\section{Results} \label{sec:results}

\subsection{Exploring LLM pre-training with nanoGPT-2 at small scale} \label{sec:results-nanogpt2}

\paragraph{LLMs trained with higher LRs observe higher gains with \method{}.} We ran controlled experiments to better understand the correlation between LR and gains due to checkpoint averaging. Initially we train nanoGPT-2 small with two different LRs $(\mathsf{6 \times 10^{-3}}, \mathsf{6 \times 10^{-4}})$, keeping batch size and all relevant hyperparameters the same.  $\mathsf{6 \times 10^{-4}}$ is the assumed optimal LR computed using a grid search reported in \cite{liu2023sophia}. Subsequently, we pre-trained the same model using an LR of $\mathsf{6 \times 10^{-3}}$, which is tenfold higher in magnitude than the former. As shown in Figure \ref{fig:nanogpt2_lr}(a), models trained with higher LR observe higher gains compared to its counterpart trained with lower LR due to post hoc checkpoint averaging in \method{}. From Figure \ref{fig:nanogpt2_lr}(b) we observe that the model trained with a higher LR converges faster but compromises on generalization, a phenomenon also observed by \cite{kaur2022maximum}. The gap in generalization is effectively mitigated by checkpoint averaging through \method{}, as shown in Figure \ref{fig:nanogpt2_lr}(c). Interestingly, we note that the training trajectory of \method{} approximates that of a model trained with a lower LR. This is an important insight: checkpoint averaging acts as a surrogate for LR decay, thereby enabling the model to be trained with a higher LR. In practical LLM pre-training scenarios, where conducting a grid search is challenging due to the model's size, adopting our proposed training recipe could be advantageous. One might select a higher LR (that doesn’t cause divergence) and train an LLM faster without compromising much generalization compared to conventional pre-training strategy.

\begin{table}[t]
  \centering
  \setlength{\tabcolsep}{3pt}
  \resizebox{0.9\columnwidth}{!}{
  \begin{tabular}{lclcccccccccc}
    \toprule
    \multirow{2}{*}{Models} & \multicolumn{2}{c}{Steps} & \multicolumn{2}{c}{Lambada openai} &  \multicolumn{2}{c}{ARC-easy} & \multicolumn{2}{c}{SciQ} & \multicolumn{2}{c}{BoolQ} & \multicolumn{2}{c}{Average} \\
    \cmidrule(lr){2-3} \cmidrule(lr){4-5} \cmidrule(lr){6-7} \cmidrule(lr){8-9} \cmidrule(lr){10-11} \cmidrule(lr){12-13}
     & &    & LAWA & Original & LAWA & Original  & LAWA & Original   & LAWA & Original &  LAWA & Original   \\ 
    \midrule
    nanoGPT-2(125M)
    & & 50 K  & 33.77 & 27.65  & 44.65 & 44.07    & 74.10 & 71.60     & 52.08 & 48.26 &  51.15 & 47.89 \\
    & & 70 K  & 35.57 & 33.57  & 44.65 & 44.36    & 75.8 & 74.6       & 54.31 & 54.43 &  52.5 & 51.74 \\
    \midrule
    nanoGPT-2(335M)
    & & 50 K  & 35.88 & 30.95  & 44.74 & 44.28    & 75.8 & 72.9     & 50.76 & 50.49 &  49.72 & 51.72 \\
    & & 70 K  & 36.95 & 33.59  & 45.96 & 45.33    & 75.2 & 74.2     & 52.94 & 51.62 &  52.76 & 51.18 \\
    \midrule
    nanoGPT-2(770M)
    & & 50 K  & 31.28 & 28.08  & 42.21 & 40.99    & 65.2 & 65.1     & 58.17 & 51.1 &  49.21 & 46.31 \\
    & & 70 K  & 33.51 & 29.92  & 43.6 & 42.72    & 68.2 & 67.7     & 56.39 & 52.6 &  50.42 & 48.23 \\
    \bottomrule
  \end{tabular}}
  \vspace{0.5em}
  \caption{The zero-shot performance of nanoGPT-2 LLMs on academic question answering and knowledge assessment downstream tasks is improved by \method{}. The checkpoints derived using our recipe require fewer steps to reach higher zero-shot performance than the checkpoints derived using original training.}
  \label{table:nanogpt2_zeroshot}
\end{table}

\begin{table}[t]
  \vspace{8pt}
  \centering
  \scriptsize
  \setlength{\tabcolsep}{3pt} 
  \resizebox{0.9\columnwidth}{!}{
  \begin{tabular}{ccccccccccccc}
    \toprule
    \cmidrule(r){1-3}
    \multirow{1}{*}{Models} & \multicolumn{2}{c}{Steps} & \multicolumn{2}{c}{Lambada openai} &  \multicolumn{2}{c}{SciQ} & \multicolumn{2}{c}{WikiText($\downarrow$)} & \multicolumn{2}{c}{ARC-easy} \\
    \cmidrule(lr){2-3}\cmidrule(lr){4-5} \cmidrule(lr){6-7} \cmidrule(lr){8-9}\cmidrule(lr){10-11}  \\
     & &    & LAWA & Original & LAWA & Original  & LAWA & Original   & LAWA & Original &    \\ 
    
    \multirow{4}{*}{Pythia-1B}
    & & 48 K  & 50.32 & 46.85  & 84.6 & 84.3    & 18.34 & 19.33     & 54.50 & 54.25 &    \\
    & & 60 K  & 50.77 & 47.00  & 84.6 & 84.4    & 17.91 & 18.82     & 55.18   & 54.50 &    \\
    & & 105 K & 57.84 & 56.39  & 86.1 & 86.3    & 16.83 & 17.12     & 56.77 & 56.27 & \\
    & & 141 K & \textbf{58.99} & 58.68  & 86.7 & 87.6    & \textbf{16.50} & 16.71     & \textbf{58.33} & 58.16 & \\
    \midrule
    \multirow{4}{*}{Pythia-2.8B}
    & & 48 K  & 63.5 & 61.9     & 86.5 & 85.6   & 14.60  & 15.37    & 61.4 & 60.6 & \\
    & & 60 K  & 64.3 & 63.8     & 86.8 & 86.3   & 14.17  & 14.87    & 62.7 & 62.1 & \\
    & & 105 K & 64.77 & 63.14   & 87.7 & 87.4   & 12.91 & 13.08     & 63.67 & 63.04 & \\
    & & 141 K & \textbf{65.47} & 65.26   & \textbf{88.8} & 88.6   & \textbf{12.59} & 12.70     & \textbf{64.68} & 64.56 & \\
    \midrule
    \multirow{4}{*}{Pythia-6.9B} 
    & & 48 K  & 65.8 & 62.3     & 88.7 & 88.0       & 13.55 & 14.25     & 63.7  & 62.9 & \\
    & & 60 K  & 67.1 & 64.6     & 88.6 &  89.0      & 13.04 & 13.61     & 64.1 & 62.9 & \\
    & & 105 K & \textbf{68.05} & 67.78   & \textbf{91.1} & 91.2       & 11.92 & 12.07           & \textbf{67.88} & 67.08 & \\
    & & 141 K & \textbf{69.08} & 68.85   & \textbf{92.0} & 91.7       & \textbf{11.61} & 11.70  & \textbf{68.13} & 67.80 & \\
    \midrule
    \multirow{4}{*}{Pythia-12B} 
    & & 48 K  & 66.8 & 65.4     & 89.7 & 88.8   & 13.09  & 13.35    & 66.4 & 65.1 & \\
    & & 60 K  & 67.8 & 66.2     & 90.3 & 90.5   & 12.54  & 12.76    & 67.3 & 60.1 & \\
    & & 105 K & \textbf{71.06} & 70.65   & 91.6 & 91.9  & 11.17 & 11.33     & 69.78 & 69.31 & \\
    & & 141 K & \textbf{71.56} & 71.00   & \textbf{92.8} & 92.3   & \textbf{10.84} & 10.91     & 70.58 & 70.74 & \\
    \bottomrule
  \end{tabular}}
  \vspace{0.5em}
  \caption{\method{} improves zero shot performance of Pythia LLMs on academic question answering and knowledge assessment downstream tasks starting very early on in the training. The checkpoints derived using \method{} requires less steps to reach higher zero-shot performance than the checkpoints derived using original training. We indicate the scores in bold font when the performance of \method{} surpasses the final score (at 141K steps) obtained using the original training or achieves comparable performance significantly earlier, specifically at 105K steps.}
  \label{table:zeroshot}
\end{table}

\begin{figure}[t]
\centering
\begin{subfigure}{.32\linewidth}
  \includegraphics[width=\linewidth, height=2.5cm]{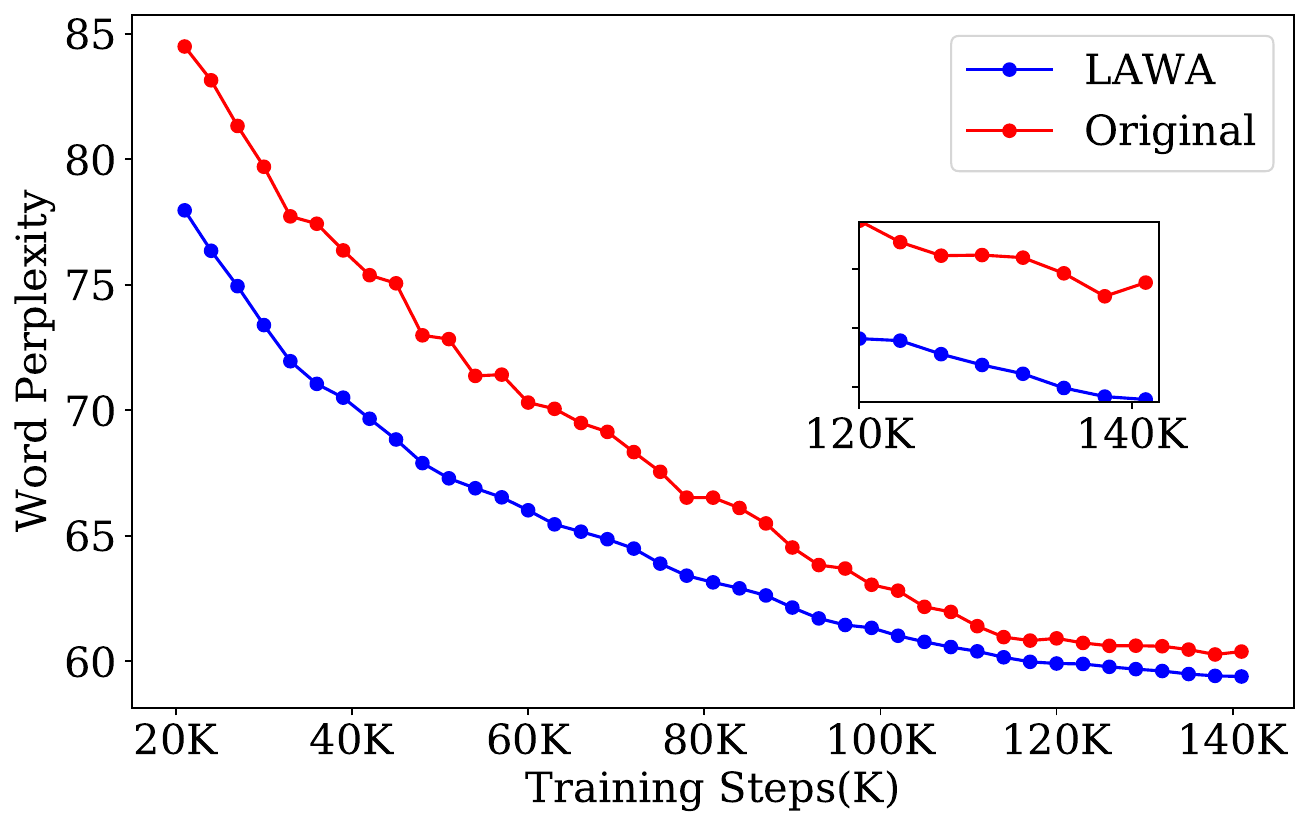}
  \caption{Philpapers}
  \label{fig:kitti_bpp_psnr}
\end{subfigure}
\begin{subfigure}{.32\linewidth}
  \includegraphics[width=\linewidth, height=2.5cm]{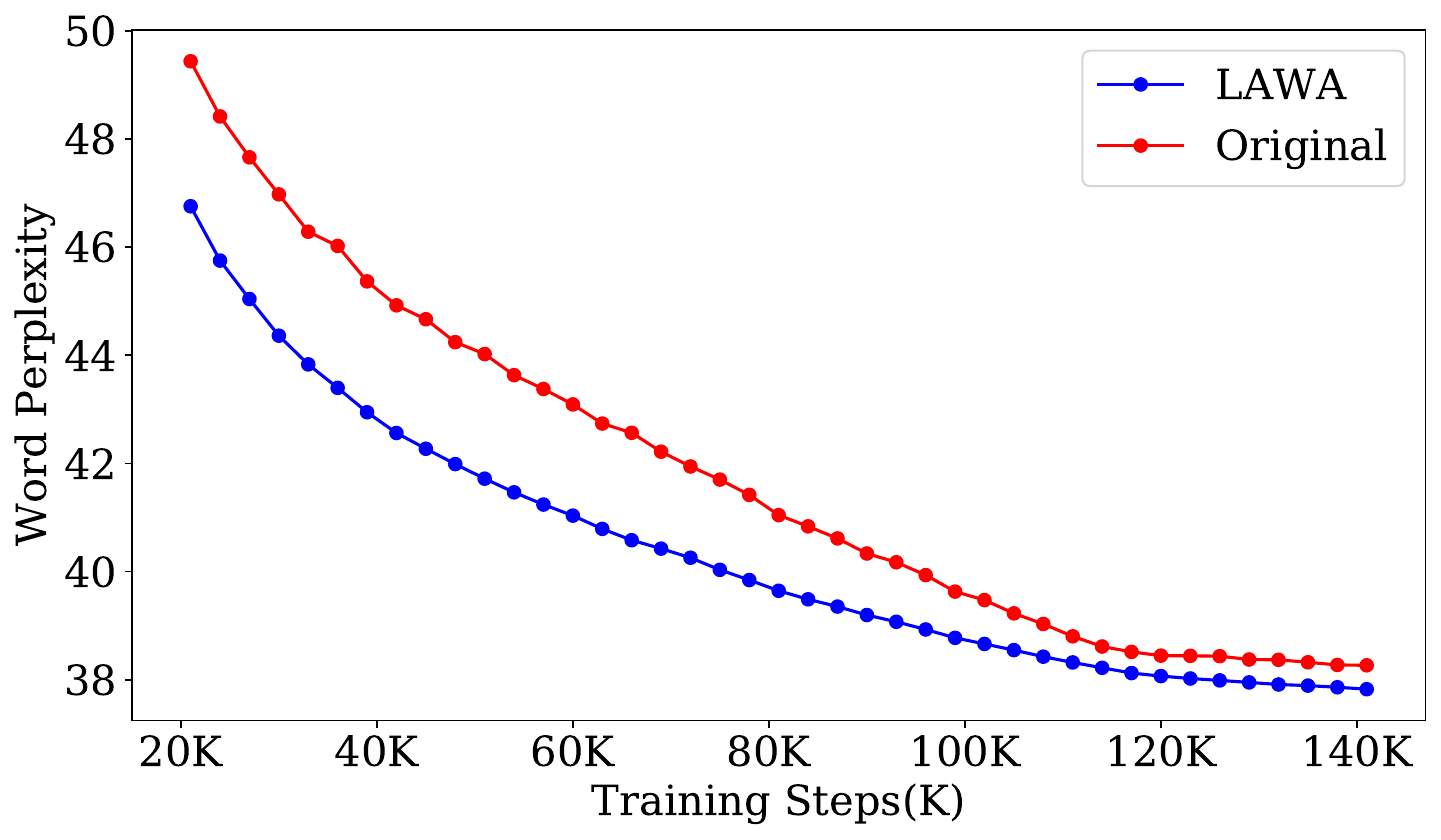}
  \caption{Bookcorpus2}
  \label{fig:kitti_bpp_ssim}
\end{subfigure}
\begin{subfigure}{.32\linewidth}
  \includegraphics[width=\linewidth, height=2.5cm]{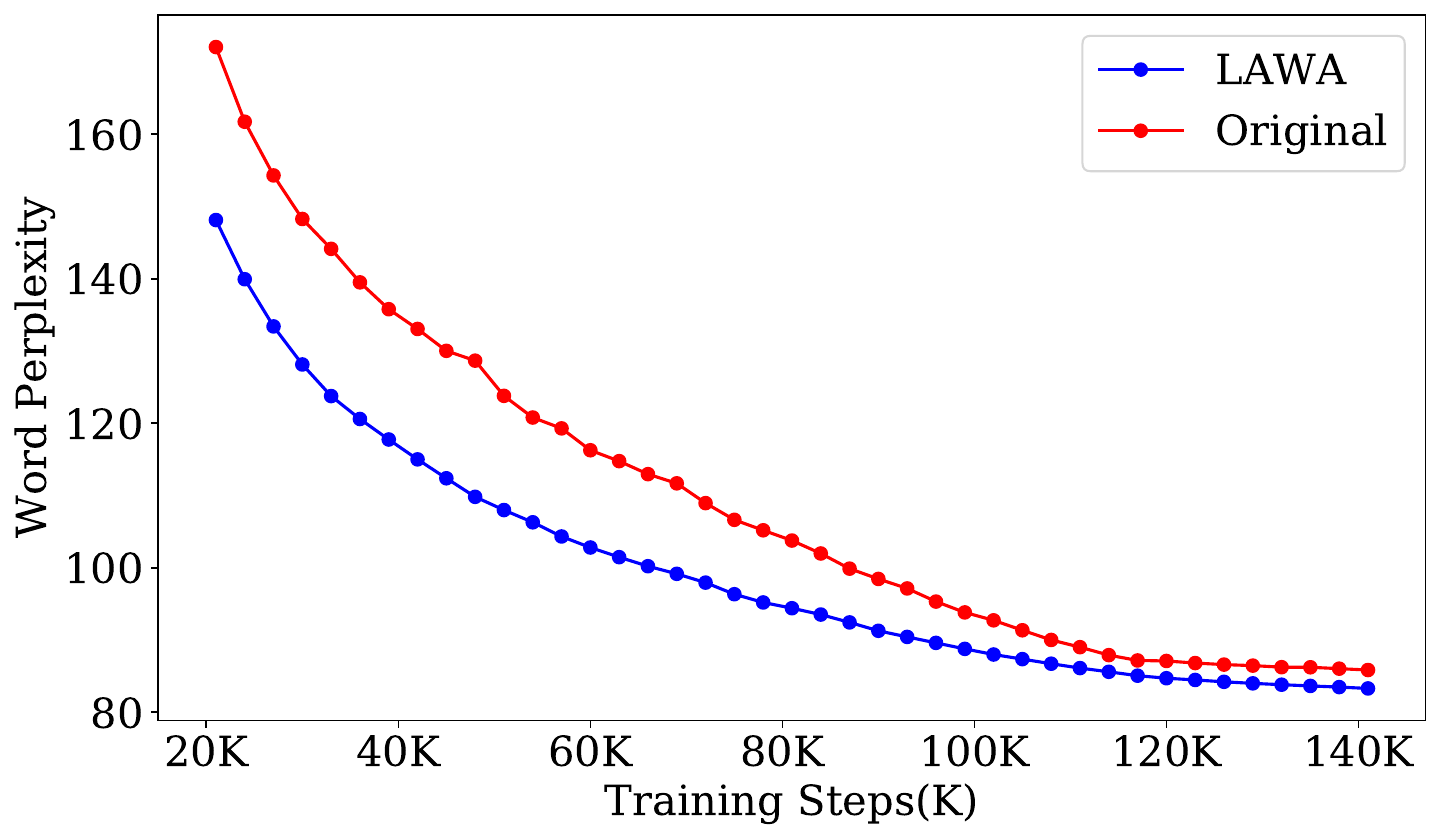}
  \caption{Youtube Subtitles}
  \label{fig:pile_youtube}
\end{subfigure}

\vspace{-0.5em}
\caption{\textbf{LAWA speeds up convergence for Pythia-1B on subset of tasks from the original pretraining dataset i.e. PILE}. We present the original and the \method{} training trajectories for 3 different tasks from PILE namely philpapers, bookcorpus2 and youtube subtitles.}
\vspace{-0.5em}
\label{fig:pile1B}
\end{figure}

\begin{figure}[t]
\centering
\begin{subfigure}{.32\linewidth}
  \includegraphics[width=\linewidth, height=2.5cm]{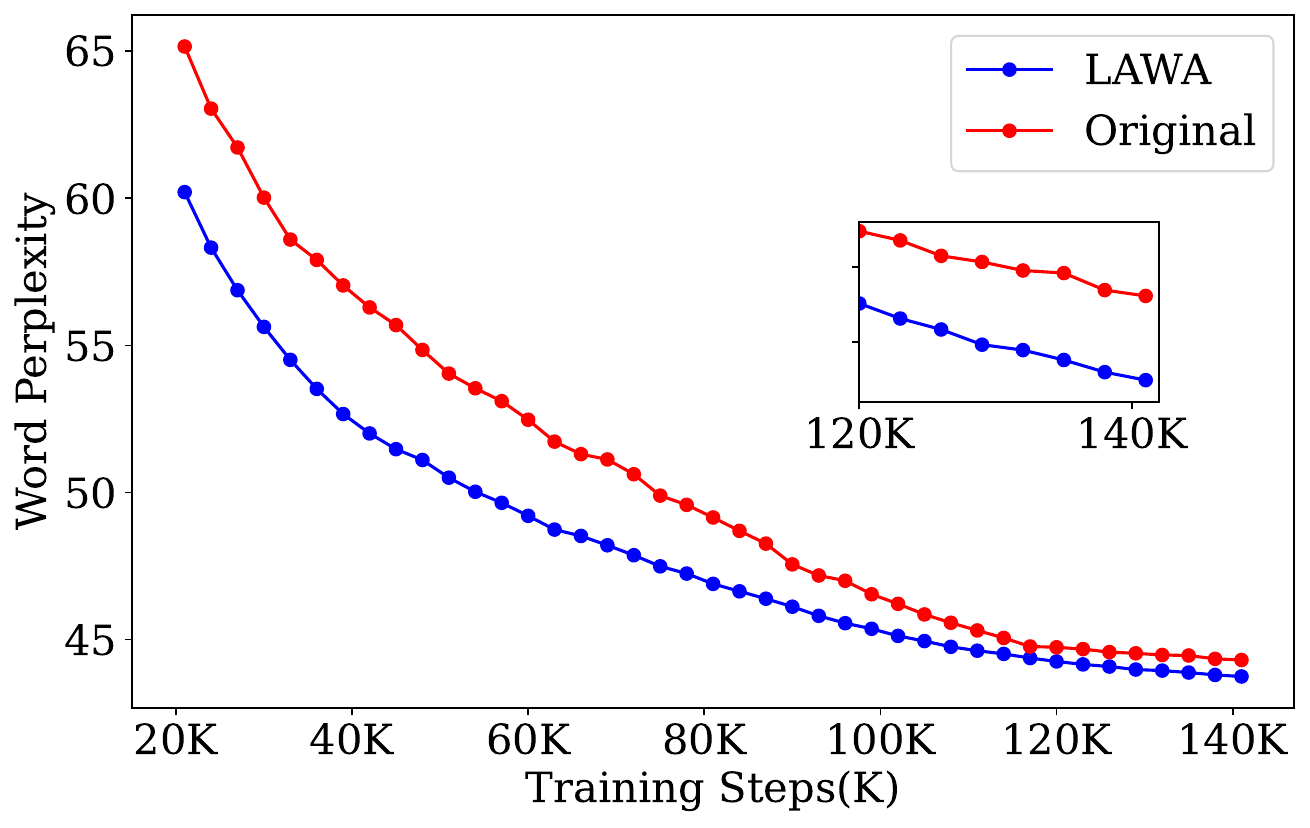}
  \caption{Philpapers}
  \label{fig:pile_philpapers}
\end{subfigure}
\begin{subfigure}{.32\linewidth}
  \includegraphics[width=\linewidth, height=2.5cm]{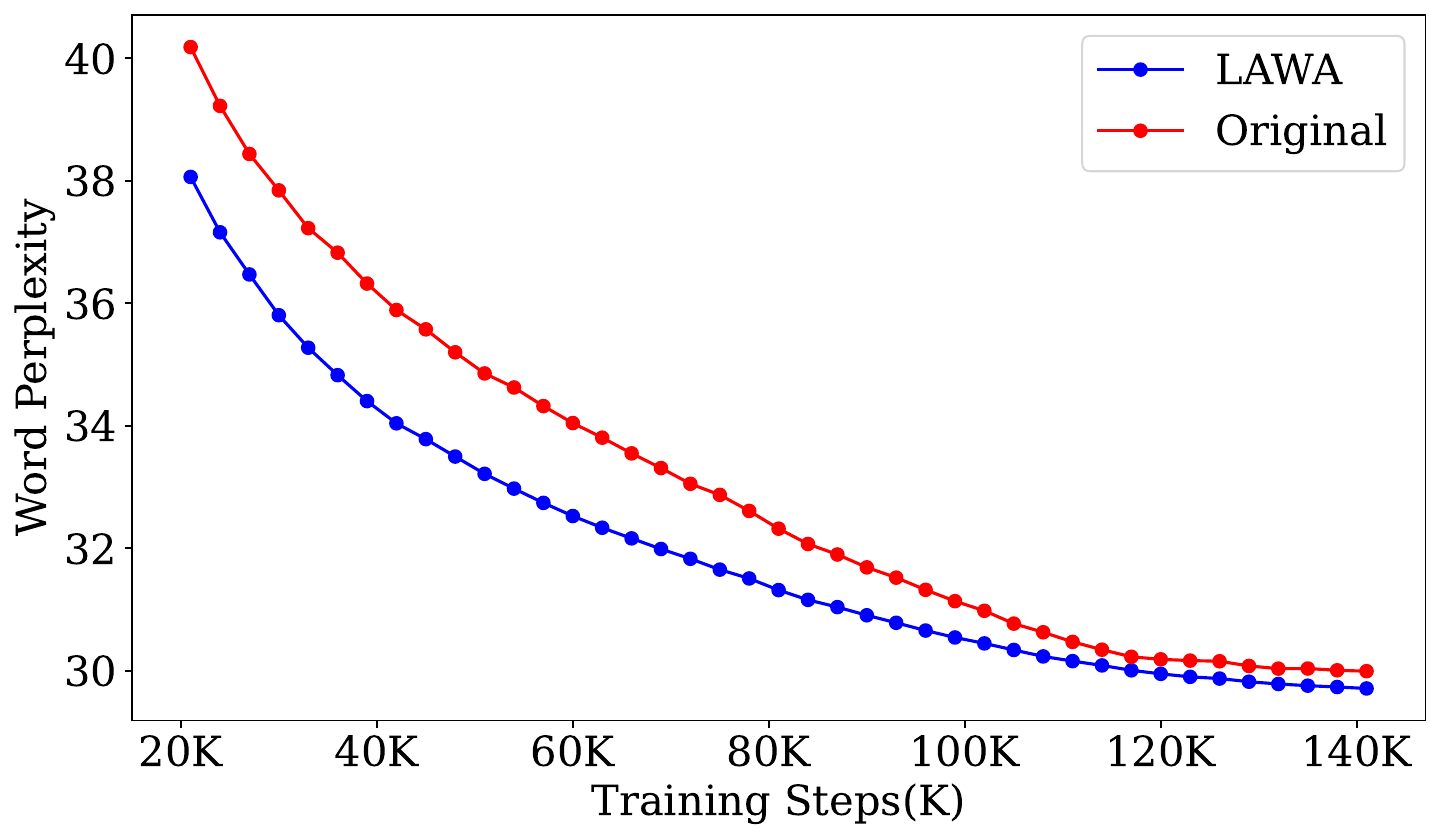}
  \caption{BookCorpus2}
  \label{fig:pile_bookcorpus2}
\end{subfigure}
\begin{subfigure}{.32\linewidth}
  \includegraphics[width=\linewidth, height=2.5cm]{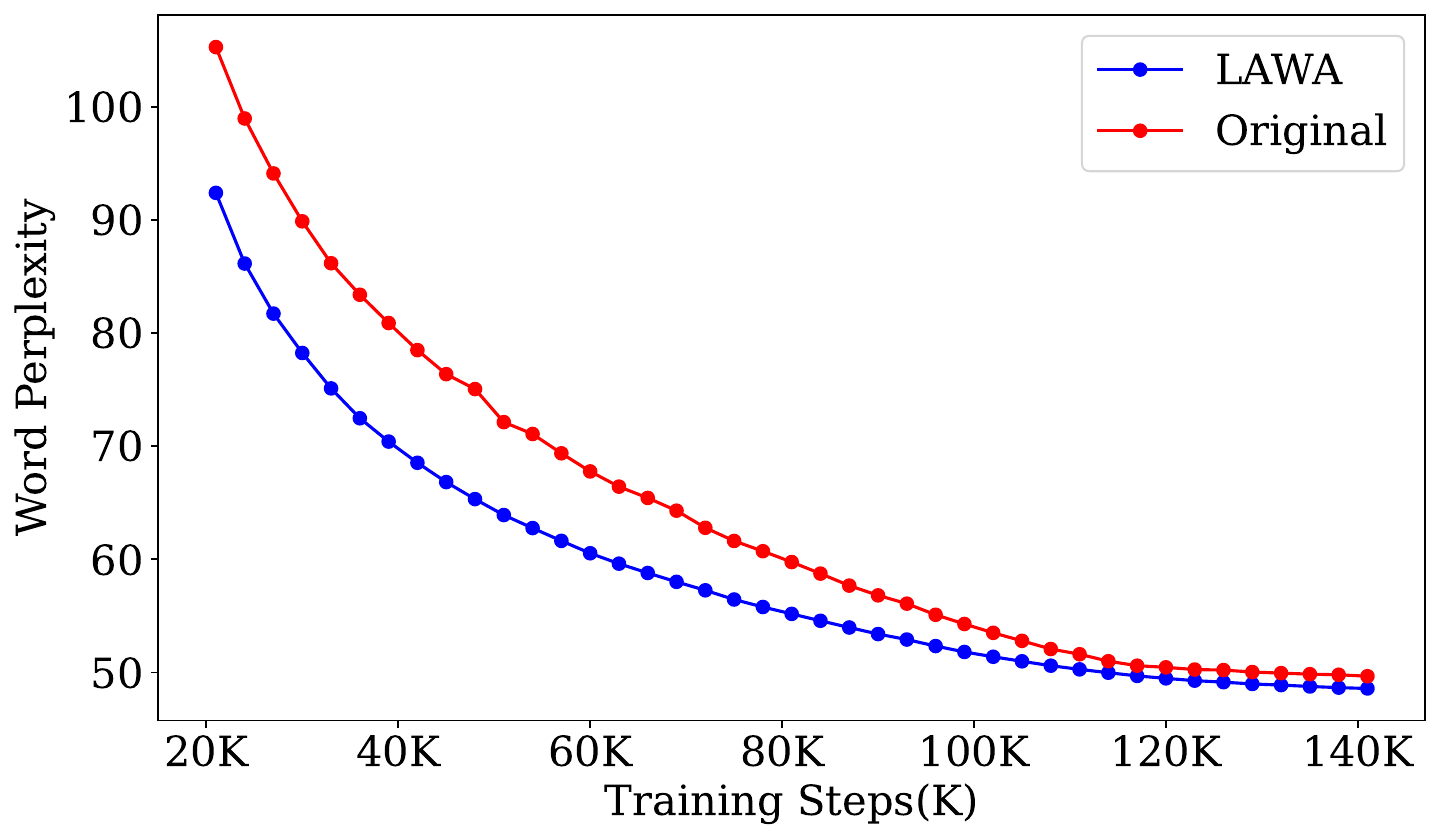}
  \caption{Youtube Subtitles}
  \label{fig:pile_youtube}
\end{subfigure}

\vspace{-0.5em}
\caption{\textbf{LAWA speeds up convergence for Pythia-2.8B on subset of tasks from the original pretraining dataset i.e. PILE.}. We present the original and the \method{} training trajectories for 3 different tasks from PILE namely philpapers, bookcorpus2 and youtube subtitles. }
\vspace{-0.5em}
\label{fig:pile2.8B}
\end{figure}

\paragraph{\method{} improves test generalization in fewer training steps compared to original pre-training and relevant baselines.} \method{} clearly outperforms the original pre-training run starting very early on during training, as shown in Figure \ref{fig:nanogpt2_main}. Since we employed a reasonably high LRs for all the nanoGPT-2 LLMs, we observe higher gains in the early-mid stages of pre-training, and the gains start diminishing towards the final stages due to the LR scheduler that continuously decays the weight throughout the training cycle.  Additionally \method{} also outperforms important baselines such as EMA and SWA. \method{} clearly has an edge over EMA throughout all training phases. Our experiments also reveal that applying SWA during the early stages of training leads to divergence (Figure \ref{fig:nanogpt2_swa_early}). Consequently, \method{} outshines SWA in both performance and ease of implementation.

\paragraph{The gains due to \method{} amplifies with far checkpoint averaging.} As shown in Figure \ref{fig:nanogpt2_main}, \method{} with higher $\mathsf{k\_stepsize}$ ($\nu$) performs better particularly for larger models. Intuitively, we believe that the diversity between nearby checkpoints might be very low given that larger models learn faster \citep{li2020train}. Hence, one needs to sample more distant checkpoints for larger models. This observation is also consistent with billion-parameter Pythia LLMs, as shown in Figure \ref{fig:pythia_farablation}.

\subsection{Scaling to Billion parameter Pythia LLMs}  \label{sec:result-1}

\begin{figure}[t]
\centering
\begin{subfigure}{.32\linewidth}
  \includegraphics[width=\linewidth, height=2.5cm]{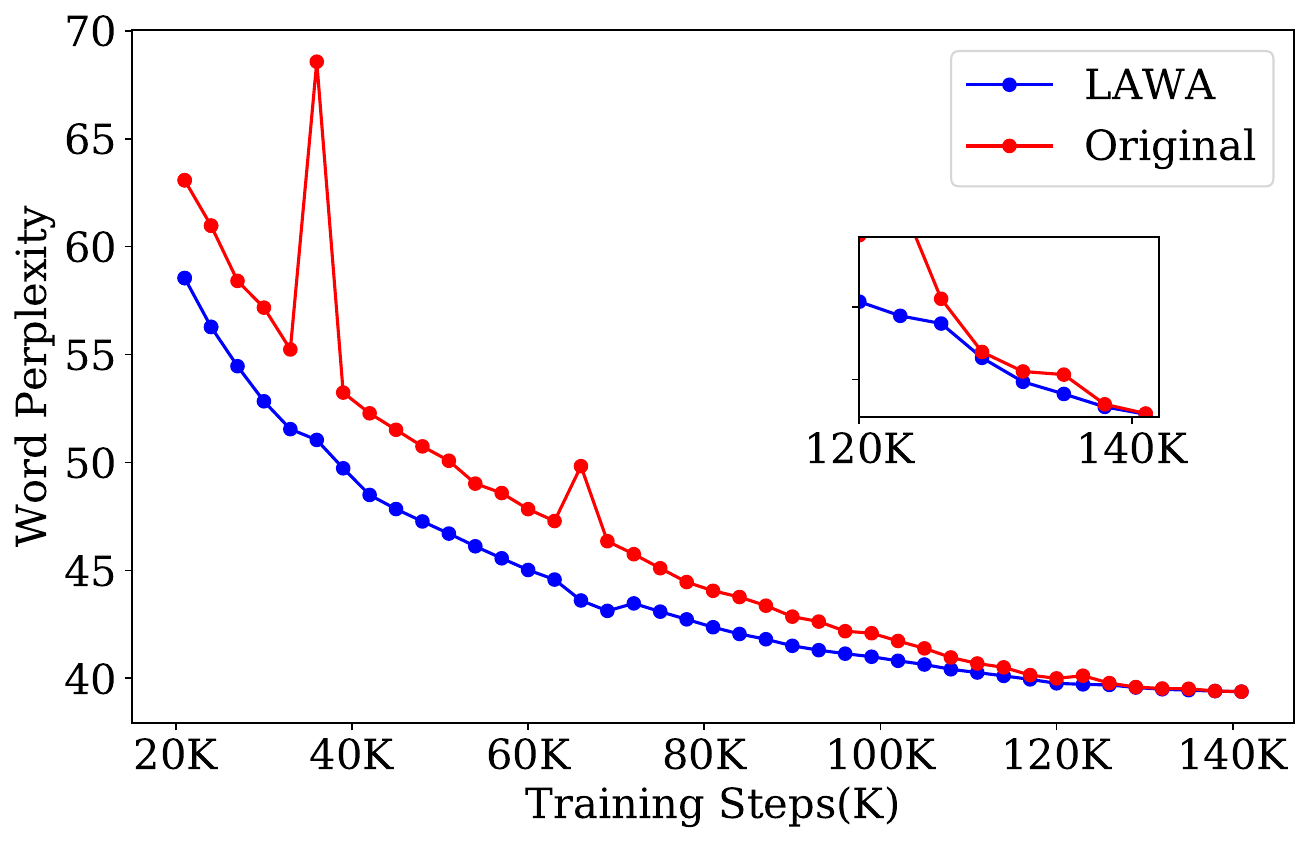}
  \caption{Philpapers}
  \label{fig:pile_philpapers}
\end{subfigure}
\begin{subfigure}{.32\linewidth}
  \includegraphics[width=\linewidth, height=2.5cm]{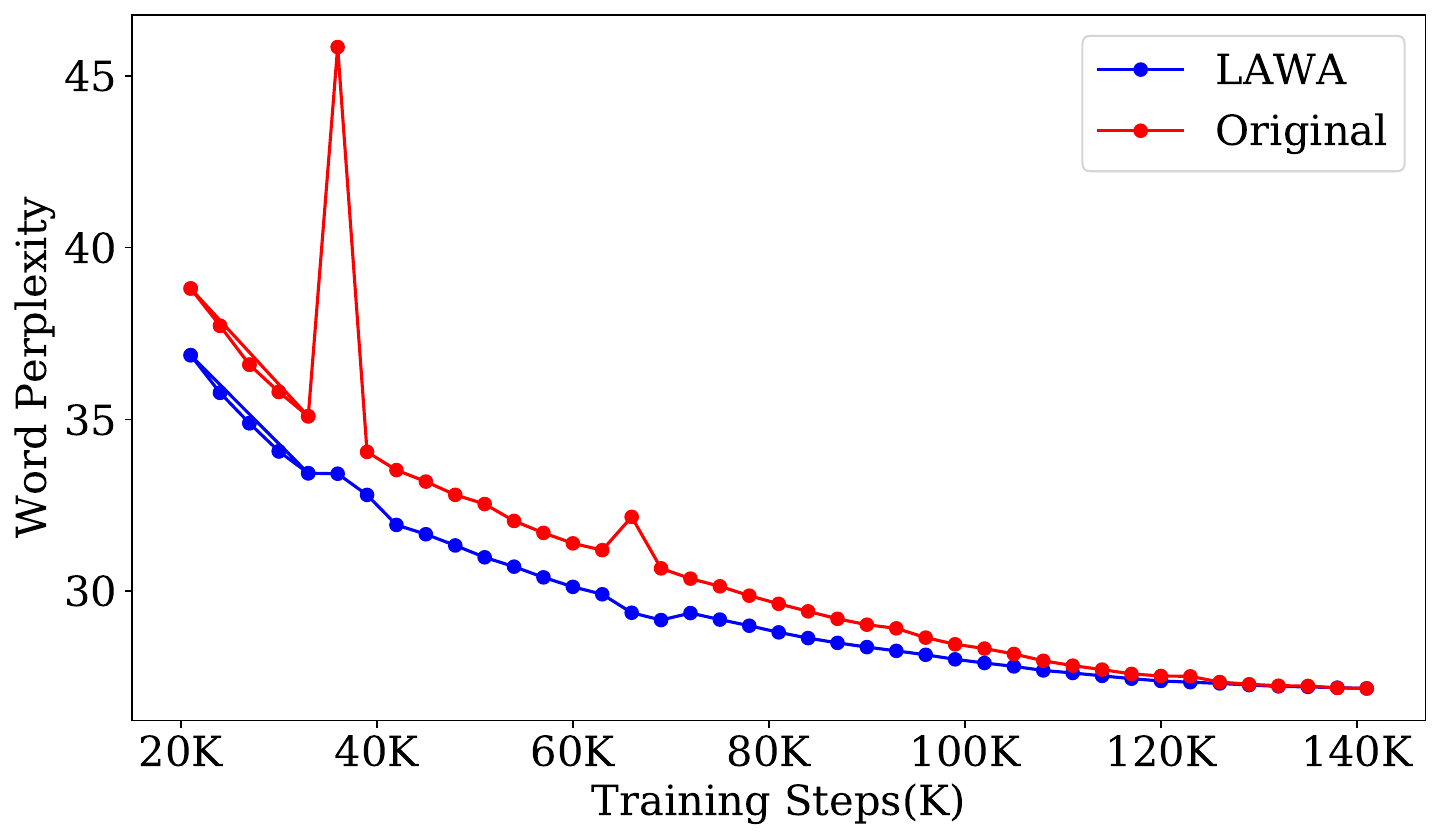}
  \caption{BookCorpus2}
  \label{fig:pile_bookcorpus2}
\end{subfigure}
\begin{subfigure}{.32\linewidth}
  \includegraphics[width=\linewidth, height=2.5cm]{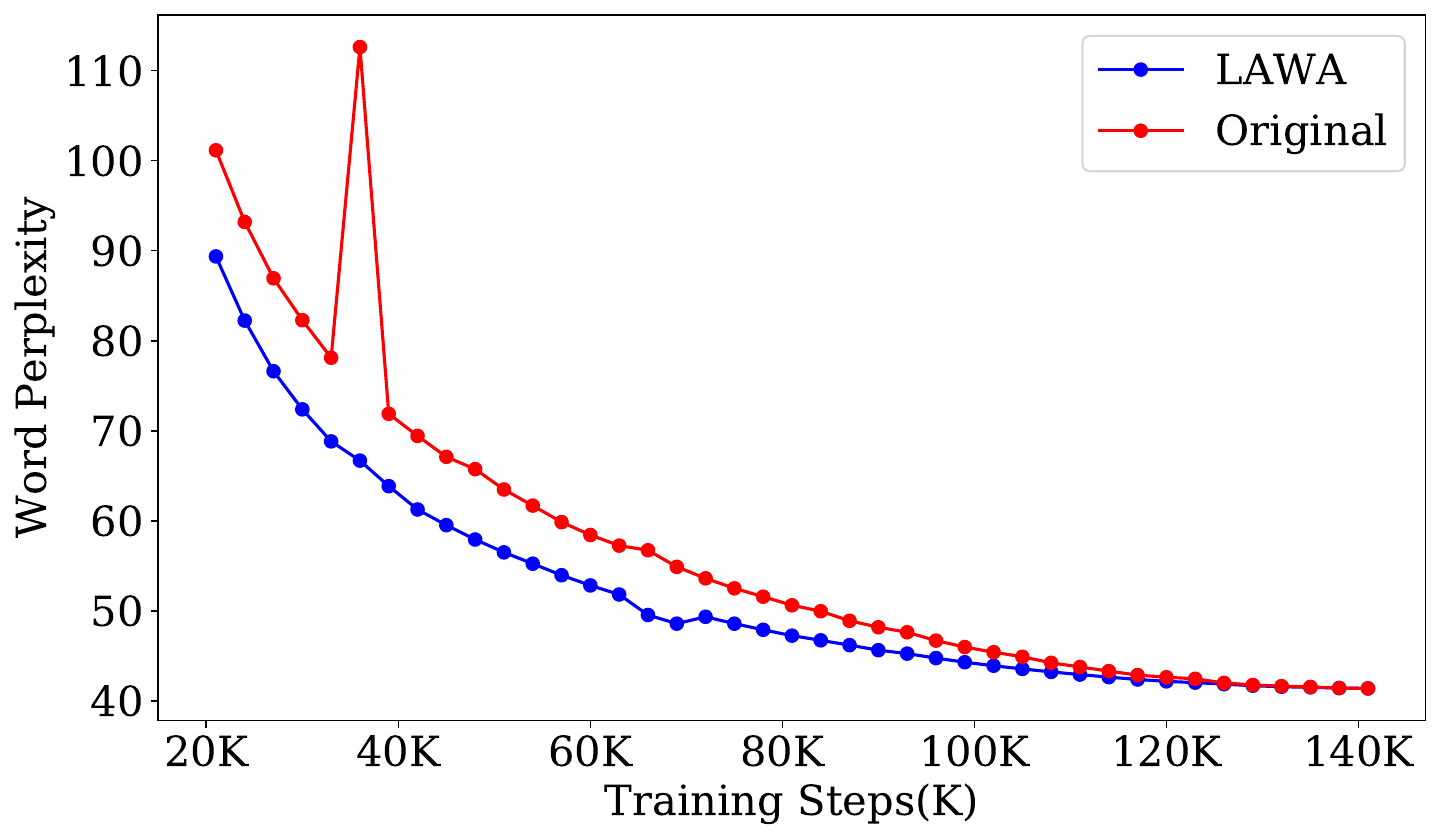}
  \caption{Youtube Subtitles}
  \label{fig:pile_youtube}
\end{subfigure}
\vspace{-0.5em}                                                                          
\caption{\textbf{LAWA speeds up convergence for Pythia-6.9B on subset of tasks from the original pretraining dataset i.e. PILE}. We present the original and the \method{} training trajectories for 3 different tasks from PILE namely philpapers, bookcorpus2 and youtube subtitles.}
\vspace{-0.5em}
\label{fig:pile6.9B}
\end{figure}

\begin{figure}[t]
\centering
\begin{subfigure}{.32\linewidth}
  \includegraphics[width=\linewidth, height=2.5cm]{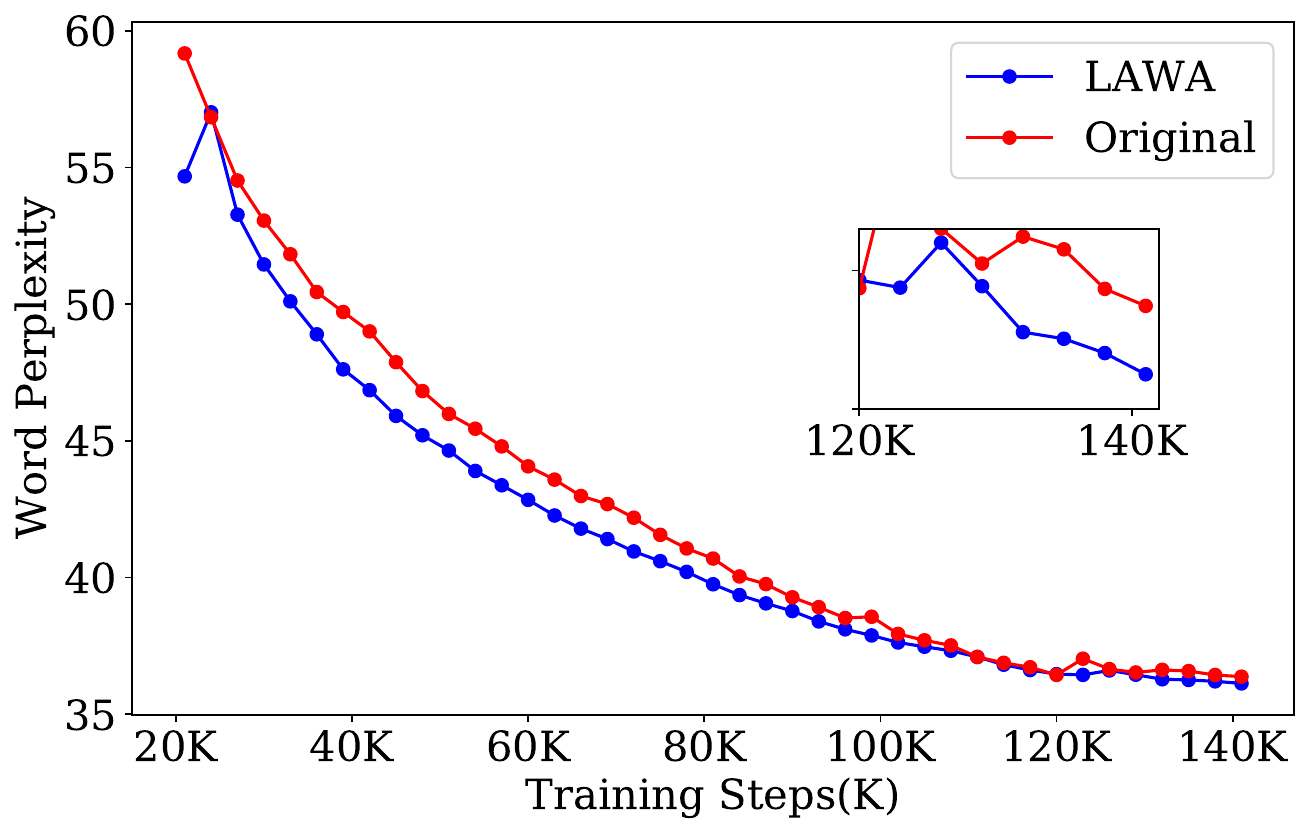}
  \caption{Philpapers}
  \label{fig:pile_philpapers}
\end{subfigure}
\begin{subfigure}{.32\linewidth}
  \includegraphics[width=\linewidth, height=2.5cm]{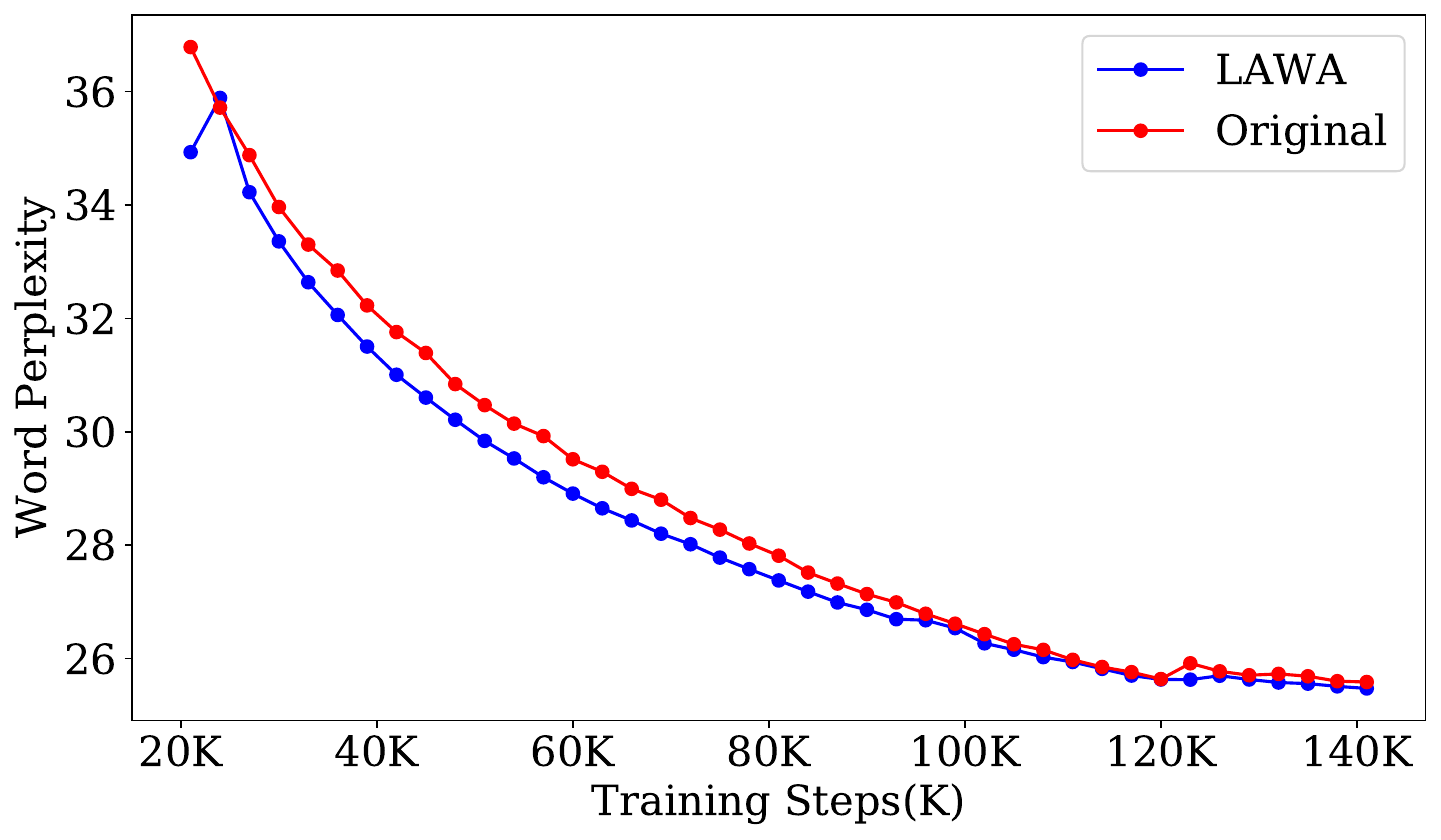}
  \caption{BookCorpus2}
  \label{fig:pile_bookcorpus2}
\end{subfigure}
\begin{subfigure}{.32\linewidth}
  \includegraphics[width=\linewidth, height=2.5cm]{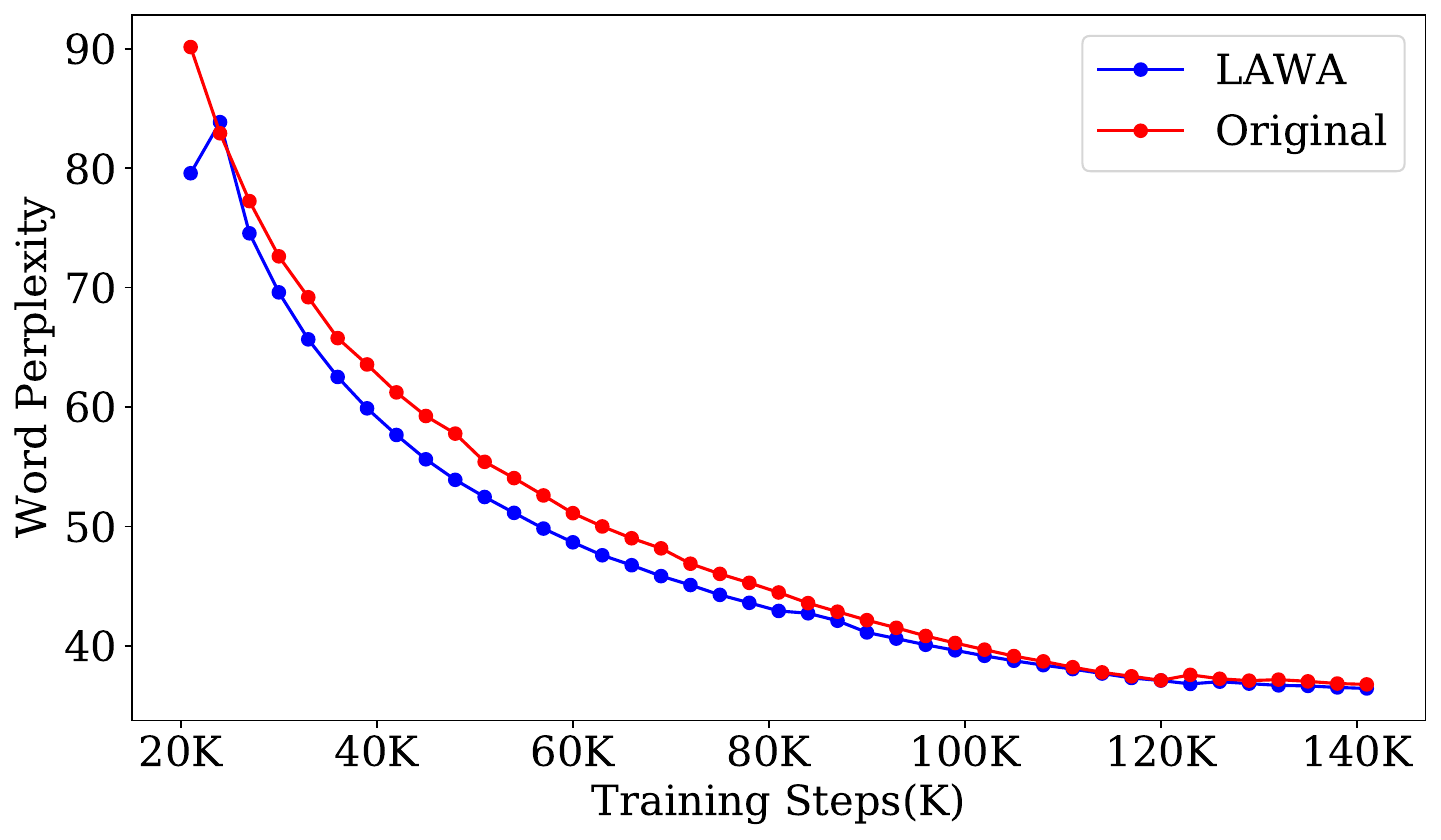}
  \caption{Youtube Subtitles}
  \label{fig:pile_youtube}
\end{subfigure}
\vspace{-0.5em}
\caption{\textbf{LAWA speeds up convergence for Pythia-12B on subset of tasks from the original pretraining dataset i.e. PILE}. We present the original and the \method{} training trajectories for 3 different tasks from PILE namely philpapers, bookcorpus2 and youtube subtitles.}
\label{fig:pile12B}
\end{figure}

Figures \ref{fig:pile1B} and \ref{fig:pile2.8B} illustrate that the checkpoints derived using \method{} demonstrate better test generalization than the checkpoints saved during original training for the Pythia-1B and Pythia-2.8B models i.e. (moderate size LLMs). In Figures \ref{fig:pile6.9B} and \ref{fig:pile12B}, we observe significant improvements in test generalization during early-mid training regime and minor improvements towards the end for both Pythia-6.9B and Pythia-12B models. All the LAWA LLMs achieve lower perplexity with lesser training steps compared to the original training trajectory, thereby saving significant amount of GPU hours (Figure \ref{fig:savings}), subsequent training costs and ingested training data. The savings are computed based on Table \ref{table:hardware} in the appendix. Additionally, \method{} proves beneficial in situations where training similar models from scratch is necessary but can only be conducted over a limited number of training steps due to strict compute budgets.

To analyze the phenomenon wherein moderate size LLMs exibit higher gains compared to their larger counterparts in test performance across both early-mid and final training trajectories, we delve into the Pythia suite's training methodologies. The authors of Pythia suite \citep{biderman2023pythia} report that they have used an exceptionally large batch size (2M tokens) and learning rate for Pythia-1B and Pythia-2.8B models to expedite the convergence. For the larger Pythia models, such as Pythia-6.9B and Pythia-12B, learning rates are reduced to $1.2 \times 10^{-4}$ while maintaining the batch sizes to 2M tokens, in line with prior work. Overall we observe thematically similar trends with the nanoGPT-2 experiments.

Recent work \citep{kaplan2020scaling,zhang2022opt,touvron2023llama,dey2023cerebrasgpt,wortsman2023stable} report \textbf{loss spikes}-- brief degradations in the performance along a training trajectory when scaling up the model size, batch size, and learning rate. In our evaluations, we observe two perplexity spikes (Figure \ref{fig:pile6.9B}). Interestingly, we find that \method{} mitigates the spikes during evaluation quite effectively. This can be intuitively explained by the \textbf{smoothing effect} -- average out the outliers to fit the trend -- resulting from averaging checkpoint weights over a range of steps that are far apart. Note that since we have sampled the checkpoints at an interval of 3K for \method{}, we may have inadvertently overlooked some checkpoints exhibiting loss spikes in models other than Pythia-6.9B. \ref{fig:pile6.9B}

\subsection{Improved Zero-shot performance}

\method{} enhances the zero-shot performance of both nanoGPT-2 and Pythia LLMs (Table \ref{table:nanogpt2_zeroshot}). In the nanoGPT-2 models, we assess zero-shot performance at 50K and 70K steps, revealing that the checkpoints generated using \method{} at 50K consistently outperform original checkpoints—derived from conventional at 50K training steps (always) and also at 70K training (in the majority of instances).

For Pythia LLMs \method{} improves the zero-shot performance in several ways. First, we observe that zero-shot performance of early-mid checkpoints (48K, 60K) achieves higher performance almost consistently, regardless of the scale as shown in Table \ref{table:zeroshot}. For instance, the \method{} Pythia-1B checkpoint evaluated at 24K steps on the Lambada OpenAI task achieves higher accuracy than the original checkpoint evaluated at the 48K step. We note that the checkpoints derived using \method{} also exhibit improvements in the later stages of training (105K,141K) on the majority of tasks, highlighted by the bolded numbers in Table \ref{table:zeroshot}. Moreover, we consistently witness gains until 105K steps across all models, which constitutes approximately 75\% of the total training steps. Therefore our recipe proves to be beneficial in a compute optimal LLM training scenario where early stopping is employed at 75\% of total training. Additionally, we find that our \method{} derived checkpoints of Pythia-6.9B reach the final accuracy/perplexity on multiple downstream tasks considerably earlier, specifically at the 105K step mark. Intuitively, we know that higher zero-shot performance on various different downstream tasks is associated with low perplexity in language modelling during training, a correlation that is mathematically substantiated \citep{saunshi2020mathematical}. Therefore, all the observations we made in Section \ref{sec:results-nanogpt2} naturally apply to zero-shot performance as well.

\begin{wrapfigure}{r}{0.5\textwidth}
    \vspace{-0.5cm}
    \centering
    \vspace{-2ex}
    \includegraphics[width=0.4\textwidth]{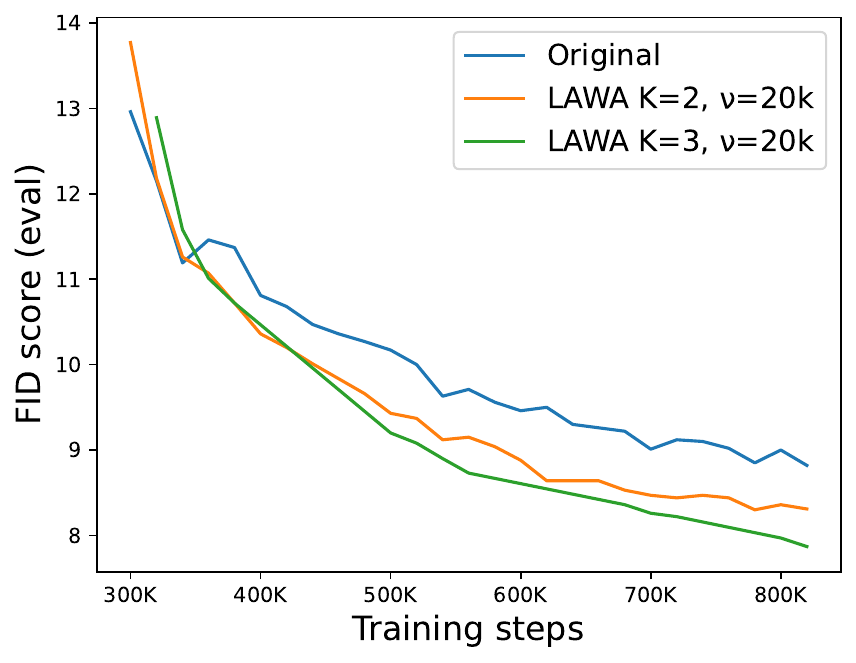}
    \caption{LAWA speeds up the convergence for Image diffusion model, measured in terms of FID on the evaluation set for ImageNet-128x128.}
    \label{fig:fid_diffusion}
\end{wrapfigure}

\subsection{Diffusion models}
We also experiment with image diffusion models to gauge the effectiveness of LAWA on generative models beyond language. The underlying model is a 422M parameter UNet \citep{ronneberger2015u,ho2020denoising} trained with $\epsilon$-prediction objective and standard cosine schedule \citep{ho2020denoising} on ImageNet 128x128 dataset. The model was trained with the Adam \citep{adam} optimizer using a learning rate of 1e-4. Figure \ref{fig:fid_diffusion} shows the FID on the evaluation set for the baseline checkpoints and LAWA averaging over the baseline checkpoints. Note that the baseline checkpoints themselves are obtained using the Exponential Moving Average (EMA) with decay rate of 0.9999 over the training trajectory, following the standard practice in training diffusion models. It is noteworthy that LAWA checkpoint averaging improves the FID over the already EMA'ed checkpoints. We defer a more thorough empirical investigation of LAWA for the family of diffusion models to future work. 


\section{Related work} \label{sec:rw}

\paragraph{Weight Averaging (WA)} has been studied and employed since the 1960s, predominantly in simple linear \citep{lakshminarayanan2018linear} and convex settings \citep{polyak_avg,neu2018iterate}. Recent WA approaches in deep learning can be broadly classified into two categories; First, approaches that simultaneously trains multiple models with different initialization and hyper-parameters \citep{model_soups,rame2022diverse, jolicoeur-martineau2023population,ramemodel} to later average them for better generalization. Second, approaches that focus on improving generalization of the final model or models close to convergence \citep{tarvainen2017mean,SWA,athiwaratkun2018improving,yang2019swalp,cha2021swad}. Stochastic WA ($\mathsf{SWA}$) \citep{SWA} employs a similar technique of averaging checkpoints along training trajectories but only works in the later stages of training (i.e. post 75\% of the training run) with a new LR scheduler. This unsual halting and restarting the training with SWA with a new schedular limits its integration. We also show that SWA when applied early on during training diverges (Section \ref{sec:abl_SWA}). Our recipe focuses on getting early gains through early averaging and can be generally applied to a wide range of training regimes. We expand the key differences of our work with SWA in Appendix \ref{sec:sup rw}.

\section{Conclusion and Future Work}

In this paper we investigated a LLM pre-training setting where the LR is significantly higher than what is conventionally used. This scenario is particularly practical as LLMs are often trained using numerous GPUs in parallel, necessitating higher batch sizes for optimal GPU utilization. Here we introduce early weight averaging throughout the training trajectory utilizing \method{}. Our findings indicate that this strategy enables LLMs to generalize more effectively in fewer steps compared to the original pre-training scheme, and key baselines as demonstrated using nanoGPT-2 and Pythia models. Looking forward we see several extensions of our work in the realm of federated fine-tuning and continual training of intermediate checkpoints.


\newpage
\bibliographystyle{iclr2024_conference}
\bibliography{want_neurips_2023}

\newpage
\appendix
\begin{center}
    \textbf{\Large \centering Supplementary Materials: Appendix}\vspace{3mm} \label{sec:appdx}
\end{center}

\section*{Contents}
\begin{itemize}
    \item \ref{sec:sup rw}: Additional Related Work.
    \item \ref{sec:sup exps}: Supplementary Experiments and Results.
    \item \ref{sec:comp}: Compute Details.
    
\end{itemize}

\section{Additional Related Work} \label{sec:sup rw}

This section serves as a supplement to the main related work detailed in Section \ref{sec:rw}. Here we further elaborate on a prior work-SWA, and highlight the key differences of \method{} in the backdrop of training LLMs. SWA \cite{SWA} is a checkpoint averaging scheme similar to \method{} but is extensively used in vision settings. In this paper we adapt SWA (also EMA) for LLM pre-training setting and presented it as a key baseline. The major differences of our recipe and SWA are listed below.

\begin{itemize}
    \item In the SWA approach the checkpoints are averaged post 75\% of regular training, i.e. close to convergence. This phase of training is termed as SWA training. Moreover the gain in generalization through SWA method is achieved at the end of training. In contrast, \method{} averages checkpoints very early on during training (post 10-15\% of training) and achieves gains in test and zero-shot generalization with fewer training steps (less than 100\% TB). SWA when applied early on during training performs poorly due to divergence of the training curve as shown in Figure \ref{fig:nanogpt2_swa_early}. 

    \item SWA modifies the learning rate scheduler mid-training and requires different learning rate schedulers for different architectures. Additionally, for models employing batch normalization, it necessitates a full pass over the entire dataset to update batch normalization statistics. \method{} is free from these constraints and is simpler to implement for large scale training.

\end{itemize}

\section{Supplementary Experiments and Results} \label{sec:sup exps}

\begin{figure}[h]
\centering
\begin{subfigure}{.45\linewidth}
  \includegraphics[width=\linewidth, height=4cm]{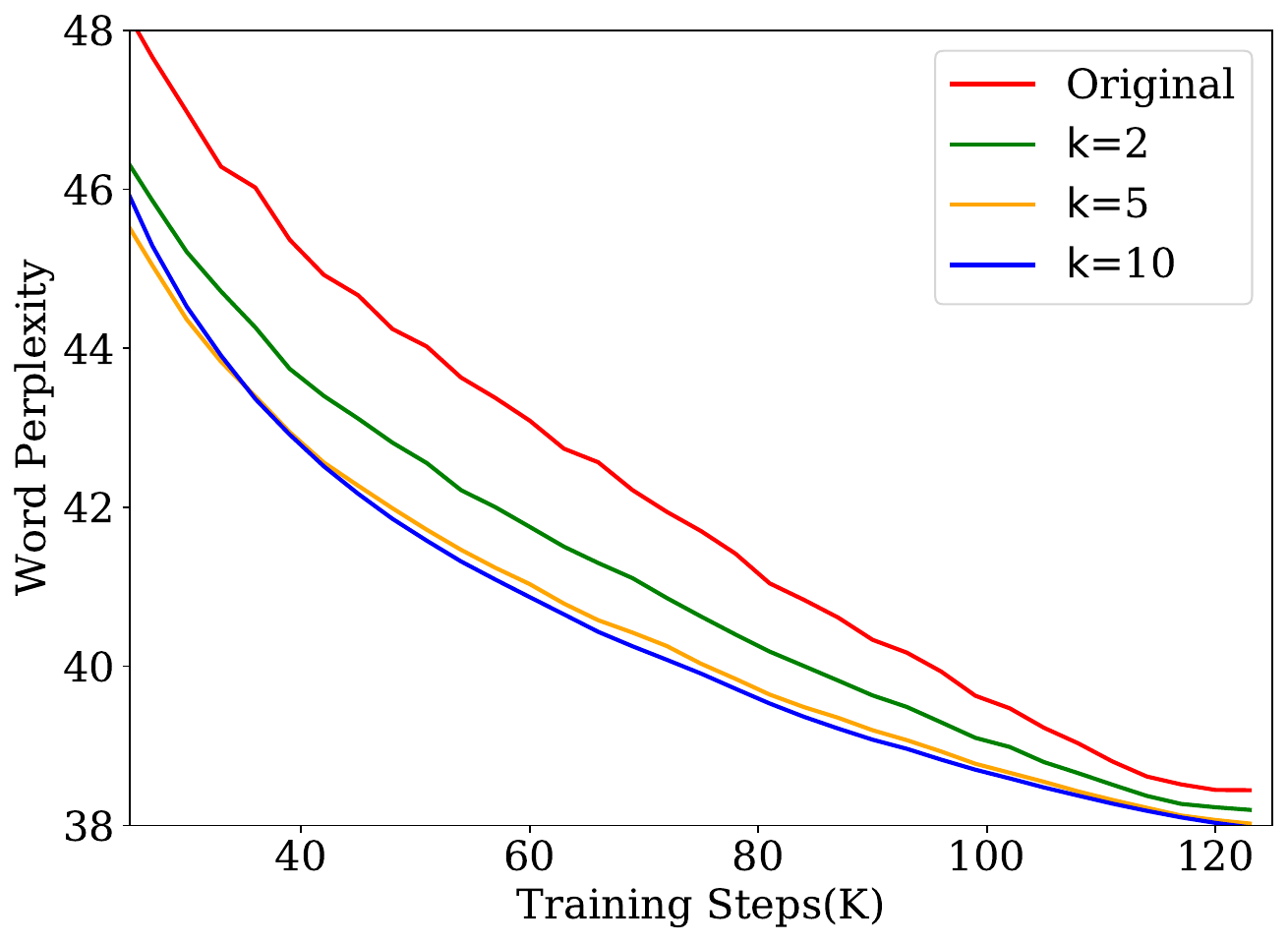}
  \caption{}
\end{subfigure}
\begin{subfigure}{.45\linewidth}
  \includegraphics[width=\linewidth, height=4cm]{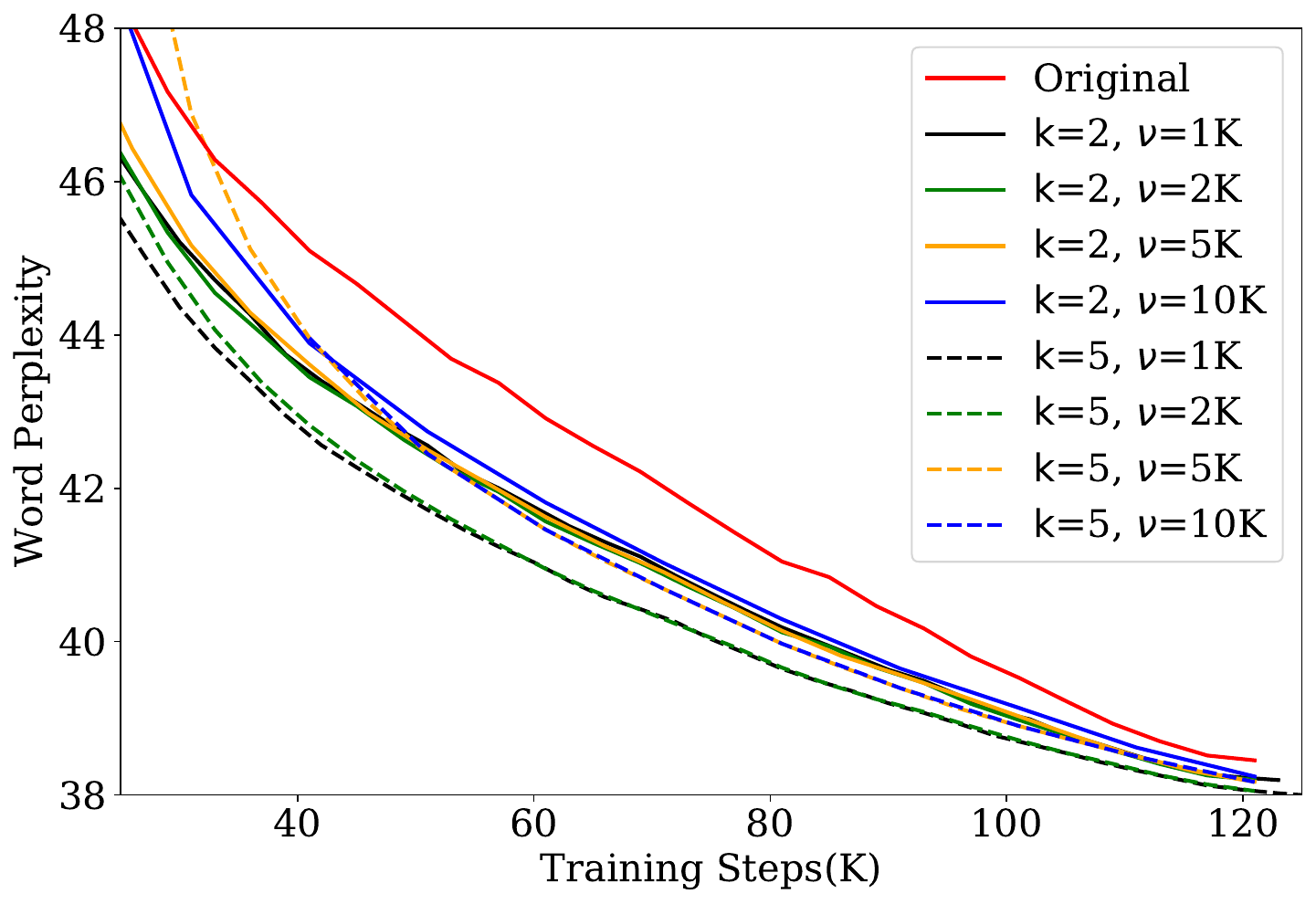}
  \caption{}
  \label{fig:pile_bookcorpus2}
\end{subfigure}
\vspace{-0.5em}
\caption{Ablations studying test performance as a function of (a) number of checkpoints to be averaged $\mathsf{k}=\{\text{2, 5, 10}\}$ at $\nu=\text{1K}$,
(b) distance of checkpoints to be averaged $\nu=\{\text{1K, 2K, 5K, 10K}\}$ at $\mathsf{k}=\{2,5\}$. The checkpoints used for ablations is Pythia-1B.}
\vspace{-0.5em}
\label{fig:pythia_farablation}
\end{figure}

\subsection{Ablations} \label{sec:abl}

We perform ablations to better understand the interplay between the performance of number of checkpoints to be averaged $\lawasol_t$ with varying $\mathsf{k}$ and distance between averaged checkpoints $\nu$ in training. We study the Pythia-1B model with held-out subset of PILE-bookcorpus2. Additionally, we also provide training trajectory of Pythia-1B model on a subset of PILE datasets much earlier than 21K steps (refer Section \ref{sec:sup exps}). 

\paragraph{Test Performance with varying $\mathsf{k}$ and fixed $\nu=1\texttt{K}$.} We investigate the impact of varying $\mathsf{k}$ on the model's test performance, while keeping $\nu$ constant at 1K. Our aim is to determine the optimal number of checkpoints to include in the average when selecting the latest $\mathsf{k}$ checkpoints following the \method{} approach. As outlined in Section \ref{sec:exp}, the Pythia checkpoints are saved at a frequency of 1K, so we have maintained $\nu$ at 1K for this analysis. From Figure \ref{fig:pythia_farablation}(a), it is apparent that a smaller $\mathsf{k}$ could be detrimental, but performance remains fairly stable for reasonably large $\mathsf{k}$ values. Consequently, for our LLM experiments, we opted for $\mathsf{k}=5$ since $\mathsf{k}=10$ tends to occupy a substantial amount of disk space, especially for larger models such as Pythia-12B.

\paragraph{Test Performance with varying $\nu$ for $\mathsf{k}=\{2,5\}$.} Memory requirements remain a key bottleneck in saving model checkpoints throughout the training trajectory particularly for extremely large billion parameter models. Therefore, it is very interesting to know how far away checkpoints in a training trajectory can be averaged? We investigate this question with far checkpoint averaging at $\nu=\{\text{1K, 2K, 5K, 10K}\}$ training steps apart for $\mathsf{k}=\{2,5\}$. As shown in Figure \ref{fig:pythia_farablation}(b), we observe for both $\mathsf{k}=2$ and $\mathsf{k}=5$, we find that averaging more recent checkpoints (keeping $\nu$ small) works better than averaging stale weights (higher $\nu$). For instance \method{} using $\mathsf{k}=5$ and $\nu=\{\text{1K, 2K}\}$ consistently performed better than against \method{} with other parameters. Overall, we empirically find that a moderate number of checkpoints ($\mathsf{k}=5$) saved in smaller frequencies ($\nu=\text{1K}$) works best.

\subsection{Early-SWA Experiments} \label{sec:abl_SWA}
\begin{figure}[t]
  \begin{subfigure}{0.33\textwidth} 
    \centering
    \includegraphics[width=\linewidth]{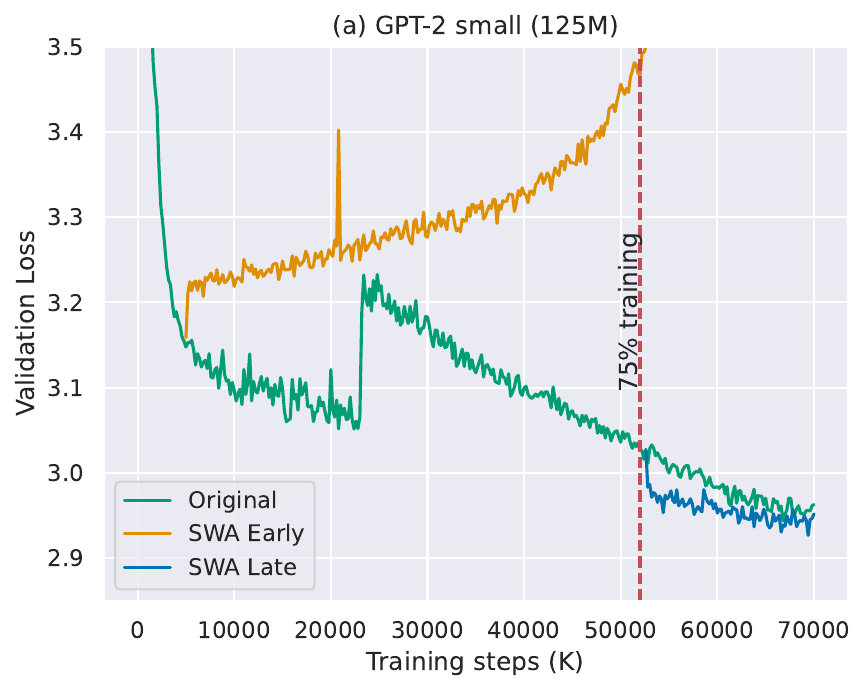} 
    \caption{}
  \end{subfigure}
  \begin{subfigure}{0.33\textwidth}
    \centering
    \includegraphics[width=\linewidth]{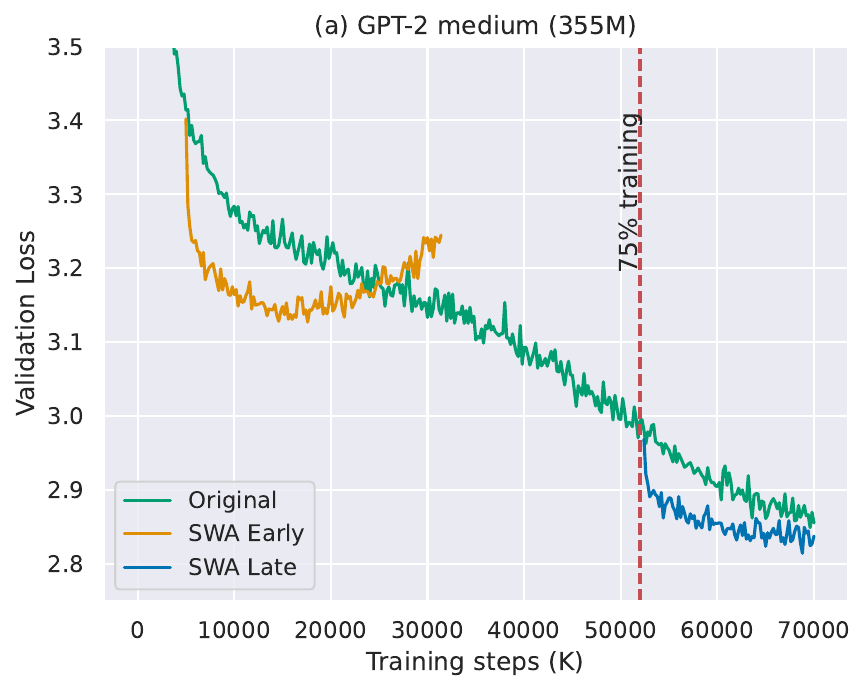} 
    \caption{}
  \end{subfigure}
  \begin{subfigure}{0.33\textwidth}
    \centering
    \includegraphics[width=\linewidth]{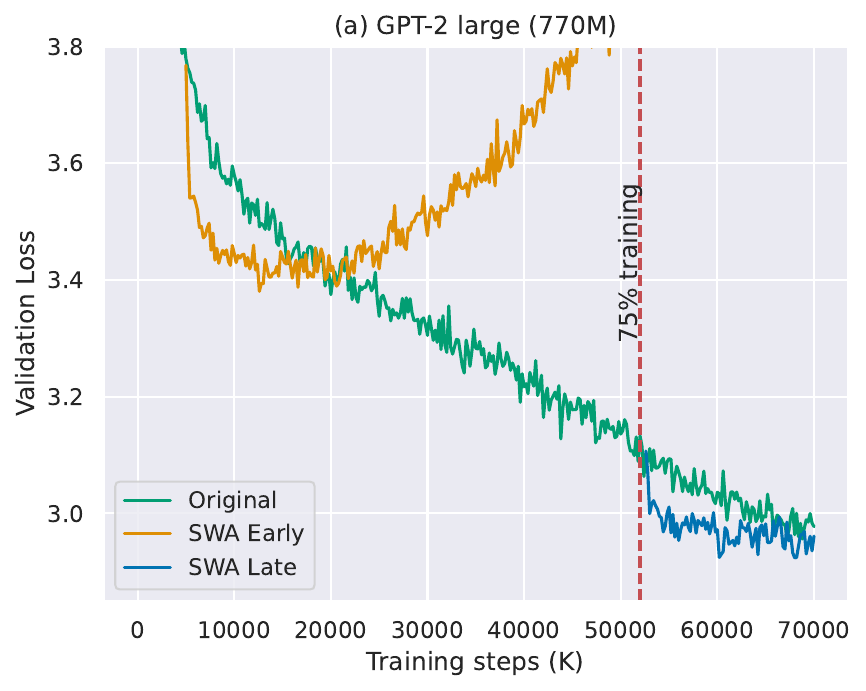} 
    \caption{}
  \end{subfigure}
  \caption{Across all model sizes (125M, 355M, 770M), we observe that the early version of stochastic weight averaging diverges, in contrast to the originally proposed late version.}
  \label{fig:nanogpt2_swa_early}
\end{figure}

Stochastic Weight Averaging \cite{SWA} has previously shown gains for smaller models, particularly in late stages of training (typically $>75\%$). We experimented with initializing SWA in early stages of training. As shown in Figure \ref{fig:nanogpt2_swa_early}, we observe that it diverges quickly - showing that SWA does not provide any gains earlier in training.

\begin{algorithm*}[ht]
\caption{Pytorch-style pseudocode of EMA/SWA}
\label{alg:ema_swa_algo}

\begin{lstlisting}[language=Python, backgroundcolor=\color{black!5}, basicstyle=\small\ttfamily, keywordstyle=\color{blue}\ttfamily, commentstyle=\color{green!40!black}\ttfamily]
def EMA_SWA(ckpt, alpha, step_size, init_point):
    """
    ckpt: list of data points (could be checkpoints or any data series)
    alpha: smoothing factor, 0 < alpha <= 1. If alpha < 0, enables SWA.
    step_size: How often to calculate average. Typically set to 1 for EMA.
    init_point: Step after which to start averaging. Typically 0 for EMA.
    """
    # Initialize the series with the first data point
    series = [ckpt[0]]
    n_models = 1
    for i in range(1, len(ckpt)):
        # Calculate EMA/SWA
        if i%step_size==0 and i>init_point:
            if alpha < 0:
                factor = n_models/(n_models + 1)
                value = (1 - factor) * series[i-1] + factor * ckpt[i]  
                n_models += 1
            else:
                value = (1 - alpha) * series[i-1] + alpha * ckpt[i]
            series.append(value)
    
    return series
\end{lstlisting}
\end{algorithm*}

\begin{figure}[h]
\centering
\begin{subfigure}{.45\linewidth}
  \includegraphics[width=\linewidth, height=4cm]{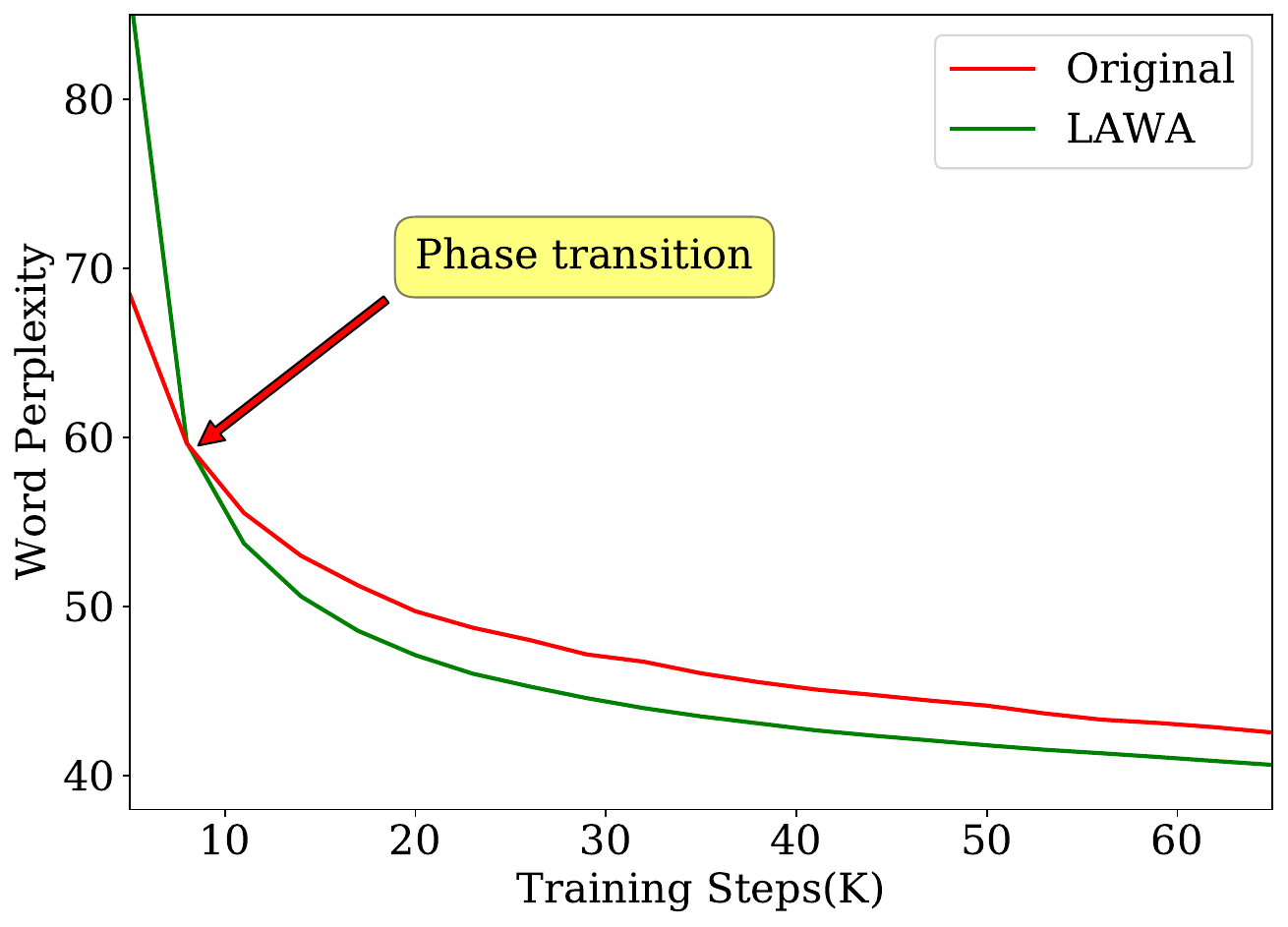}
  \caption{BookCorpus2}
  \label{fig:pile_philpapers}
\end{subfigure}
\begin{subfigure}{.45\linewidth}
  \includegraphics[width=\linewidth, height=4cm]{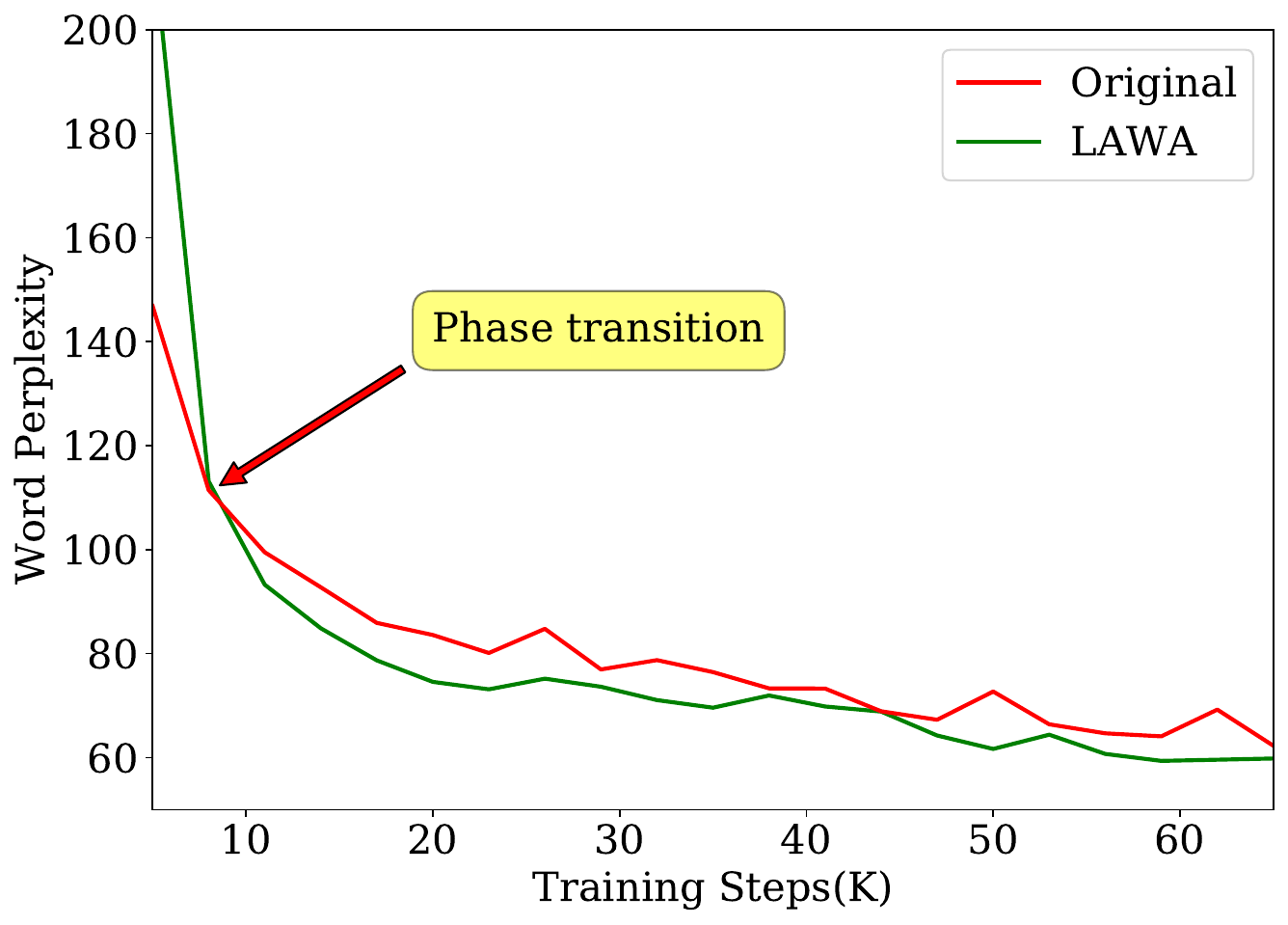}
  \caption{Enron Emails}
  \label{fig:pile_bookcorpus2}
\end{subfigure}
\caption{\textbf{Early weight averaging doesn't work at the very beginning of the training but works reasonably early during the training process.} Here we compare original and LAWA early training trajectories for Pythia-1B model on 2 different tasks namely bookcorpus2 and enron emails using held out set.}
\vspace{-0.5em}
\label{fig:supp_plots}
\end{figure}

\subsection{Phase Transition and Linear Mode Connectivity}

Averaging very initially i.e. before $~\text{8K}$ steps during training may not always yield beneficial results (Figure \ref{fig:supp_plots}). However, the technique does start showing efficacy fairly early in the training process. We highlight this phenomenon by presenting experimental results with the Pythia-1B model using a held-out set of PILE-bookcorpus2 and PILE-enron emails. We observe that \method{} trajectory undergoes a phase transition at the 8K training step. Beyond this transition, significant improvements in test performance can be seen. Such a phase transition may not occur for all Pythia LLMs. Following this phenomenon we presented our results starting 21K steps in Figures \ref{fig:pile1B}-\ref{fig:pile12B}. We further examine this phenomenon through the lens of linear mode connectivity.

\begin{figure}[t]
\begin{subfigure}{.30\linewidth}
  \includegraphics[width=\linewidth, height=3cm]{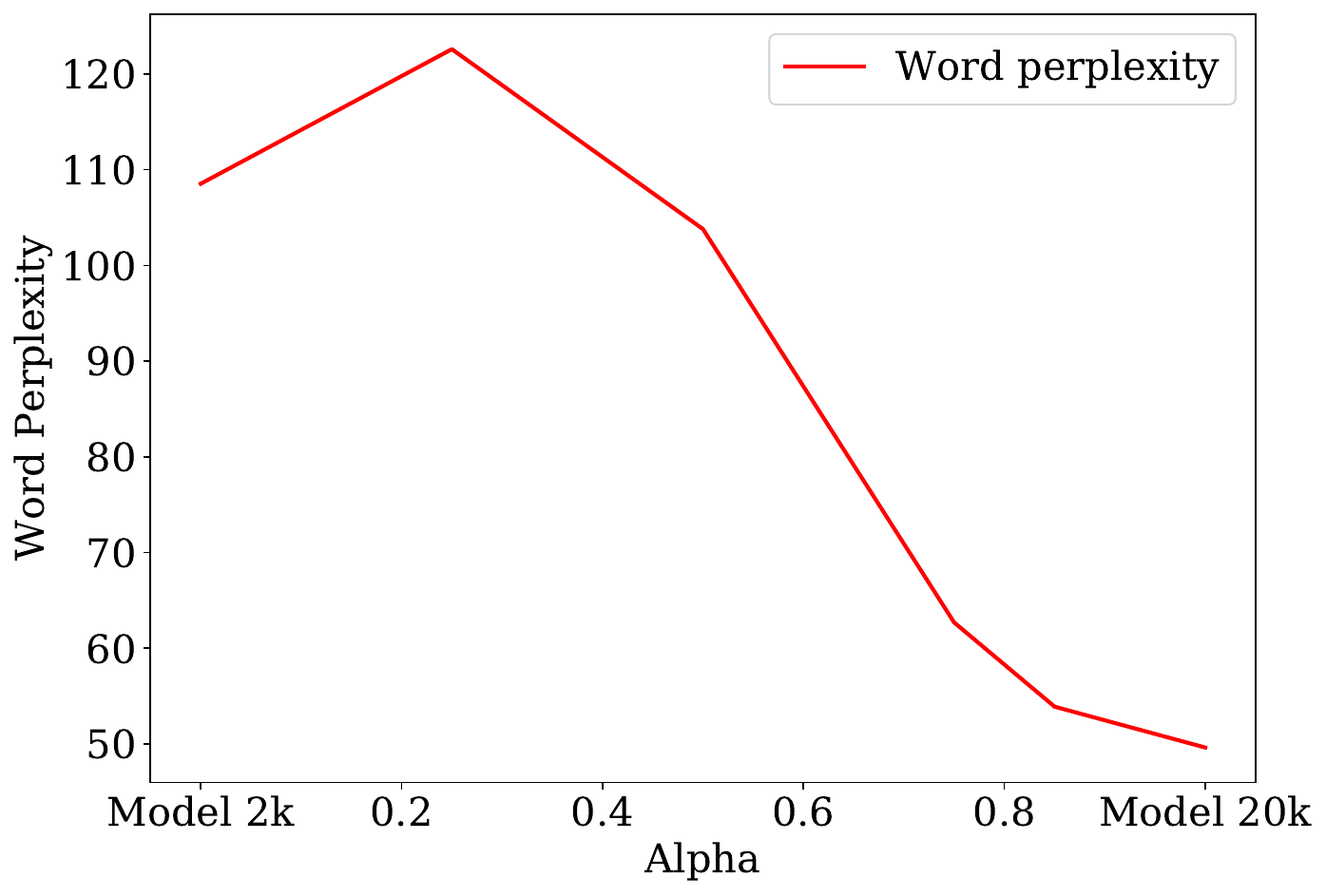}
  \caption{}
  \label{fig:pile_philpapers}
\end{subfigure}
\begin{subfigure}{.30\linewidth}
  \includegraphics[width=\linewidth, height=3cm]{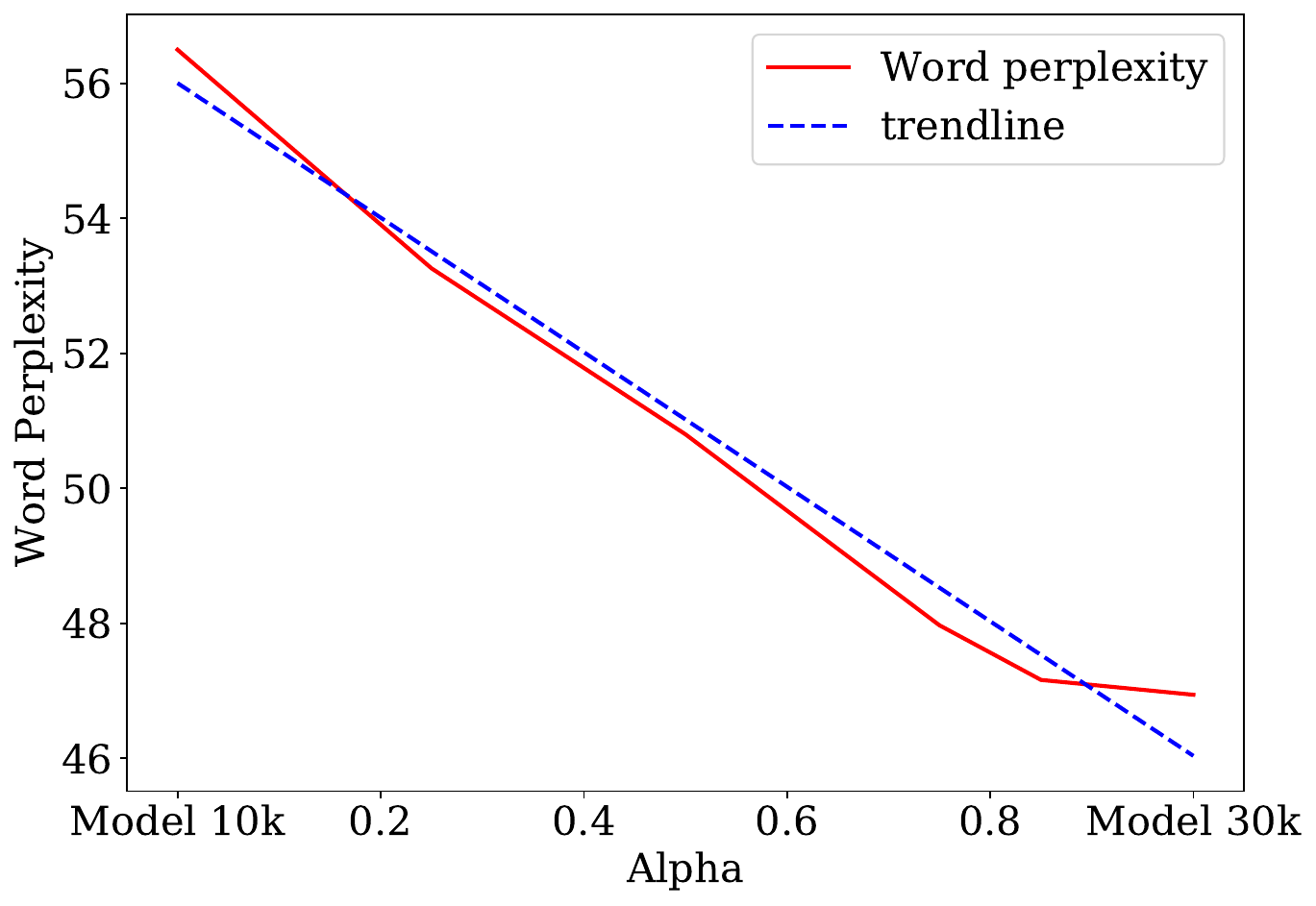}
  \caption{}
  \label{fig:pile_bookcorpus2}
\end{subfigure}
\begin{subfigure}{.30\linewidth}
  \includegraphics[width=\linewidth, height=3cm]{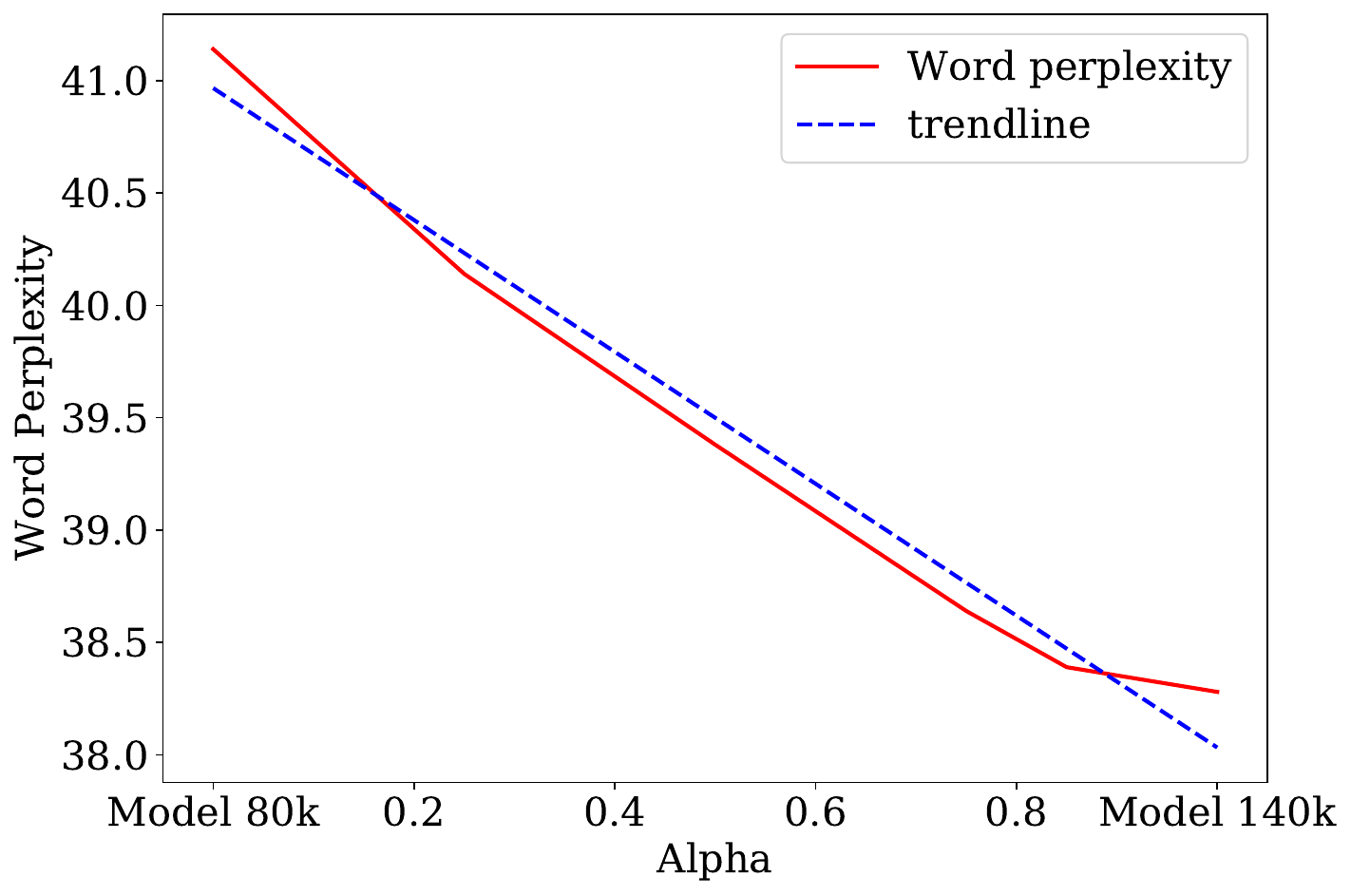}
  \caption{}
  \label{fig:pile_bookcorpus2}
\end{subfigure}
\caption{\textbf{The model checkpoints attains linear mode connectivity quite early but not at the very beginning of the training process.} We plot word perplexity as a function of the model derived from the convex combination of 2 different checkpoints i.e. $\params_{\text{LMC}}$ at $\mathsf{\alpha}=\{\text{0, 0.2, 0.4, 0.6, 0.8, 1}\}$.} In (a) we see an error barrier that means model checkpoint at 2K and 20K are not linear mode connected, whereas both (b) and (c) shows the checkpoints under consideration are linear model connected.
\label{fig:lmc}
\end{figure}

To better comprehend the linear model connectivity of checkpoints, we perform a convex combination of model checkpoints at different training stages. For instance, a model checkpoint at 2K and 20K can be combined in this manner: $\alpha \times \params_{2k} + (1 - \alpha) \times \params_{20k}$. In Fig. 10, we plot word perplexity as a function of $\alpha$ using PILE-bookcorpus2. Here we observe that initially the model checkpoints are not linear mode connected. However, based on the evaluated checkpoints shown in Figure \ref{fig:lmc}, we posit that the model checkpoint attains linear mode connectivity (LMC) quite early and maintains this property until the end of training.

\section{Amount of Compute} \label{sec:comp}
We compute the savings in GPU hours based on the Table. 6 of Pythia suite \cite{biderman2023pythia} as shown below.

\begin{table}[h]
    \centering
    \begin{tabular}{rccc}\toprule
  Model Size & GPU Count & Total GPU hours required\\\midrule
         1.0 B  & 64 & 4,830 \\
         2.8 B  & 64 & 14,240\\
         6.9 B  & 128 & 33,500 \\
         12 B   & 256 & 72,300 \\\midrule
         Total   &  & 136,070 \\\bottomrule
    \end{tabular}
    \vspace{0.5em}
    \caption{Table from \cite{biderman2023pythia}. Model sizes in the Pythia suite, number of GPUs used during training, and the total number of GPU hours, calculated as (iteration time (s) $\times$ number of iterations $\times$ number of GPUs $\div$ 3600 s/hour). All GPUs are A100s with 40GB of memory.}
	\label{table:hardware}
\end{table}

\end{document}